\documentclass[10pt,journal,compsoc]{IEEEtran}

\ifCLASSOPTIONcompsoc
  \usepackage[nocompress]{cite}
\else
  \usepackage{cite}
\fi
\ifCLASSINFOpdf
\else
\fi
\usepackage{amsmath}
\usepackage{algorithmic}

\usepackage{array}
\usepackage{algorithm}

\makeatletter
\let\NAT@parse\undefined
\makeatother
\usepackage{overpic}
\usepackage{amsfonts,amssymb}
\usepackage{subfig}
\usepackage{booktabs}
\usepackage{makecell}

\usepackage{graphicx}
\usepackage{multirow}
\usepackage{xurl}
\usepackage{wrapfig,lipsum}
\usepackage{float}
\usepackage[numbers]{natbib}
\usepackage{xcolor}
\usepackage{pifont}

\newtheorem{prop}{Proposition}

\newtheorem{corollary}{Corollary}
\newtheorem{assumption}{Assumption}
\newtheorem{definition}{Definition}

\usepackage[colorlinks,urlcolor=blue,anchorcolor=blue,citecolor=blue]{hyperref}

\begin{document}
%
\title{Influence Function Based Second-Order Channel Pruning: Evaluating True Loss Changes For Pruning Is Possible Without Retraining}

%
%
%
%

\author{Hongrong~Cheng*,~Miao Zhang*$\dagger$,~\IEEEmembership{Member,~IEEE},
Javen~Qinfeng~Shi,~\IEEEmembership{Member,~IEEE}
\IEEEcompsocitemizethanks{
\IEEEcompsocthanksitem *Equal contribution. \\
H. Cheng and J. Q. Shi are with the University of Adelaide. M. Zhang is with Harbin Institute of Technology (Shenzhen).\\
E-mail:\{hongrong.cheng, javen.shi\}@adelaide.edu.au,\\
\{zhangmiao\}@hit.edu.cn.\\
$\dagger$ Corresponding author.}
\thanks{Preprint. Under review.}}

\IEEEtitleabstractindextext{%
\begin{abstract}

Channel pruning is attracting increasing attention in the deep model compression community due to its capability of significantly reducing computation and memory footprints without special support from specific software and hardware. A challenge of channel pruning is designing efficient and effective criteria to select channels to prune. A widely used criterion is minimal performance degeneration, e.g., loss changes before and after pruning being the smallest. To accurately evaluate the truth performance degeneration requires retraining the survived weights to convergence, which is prohibitively slow. Hence existing pruning methods settle to use previous weights (without retraining) to evaluate the performance degeneration. However, we observe that the loss changes differ significantly with and without retraining. It motivates us to develop a technique to evaluate true loss changes without retraining, using which to select channels to prune with more reliability and confidence. We first derive a closed-form estimator of the true loss change per pruning mask change, using influence functions without retraining. Influence function is a classic technique from robust statistics that reveals the impacts of a training sample on the model's prediction and is repurposed by us to assess impacts on true loss changes. We then show how to assess the importance of all channels \textit{simultaneously} and develop a novel global channel pruning algorithm accordingly. We conduct extensive experiments to verify the effectiveness of the proposed algorithm, which significantly outperforms the competing channel pruning methods on both image classification and object detection tasks. One of the attractive properties of our algorithm is that it automatically obtains the prune percentage without the cumbersome yet commonly used sensitivity analysis by local pruning. To the best of our knowledge, we are the first that shows evaluating true loss changes for pruning without retraining is possible. This finding will open up opportunities for a series of new paradigms to emerge that differ from existing pruning methods. The code is available at \url{https://github.com/hrcheng1066/IFSO}. 
\end{abstract}

}

\maketitle

\IEEEdisplaynontitleabstractindextext

%
\IEEEpeerreviewmaketitle

\ifCLASSOPTIONcompsoc
\IEEEraisesectionheading{\section{Introduction}
\label{sec:introduction}}
\else
\section{Introduction}
\label{sec:introduction}
\fi

Deep neural networks have made immense strides in many domains, such as computer vision (\cite{elkerdawy2022fire, huang2022robust, liu2022multi}), natural language processing (\cite{devlin2019bert, hernandez2022natural}), and speech recognition (\cite{bohnstingl2021towards, ding2022audio}). Despite the remarkable success, their heavy computational and memory footprints hinder the application of intelligent edge systems \cite{elkerdawy2022fire,He2020learning,you2019gate}. For example, the embedded devices, which are usually equipped for many real-life applications (e.g., smart mining, field rescue, and bushfire prevention), have very limited computing and memory capacity. A typical neural network model can easily exceed such small devices' computational and memory constraints \cite{luo2017thinet, liu2021group}. To relieve this issue, researchers have proposed various neural network compression techniques to obtain lightweight models. These techniques include neural network pruning (\cite{liu2021group, yu2022combinatorial,nonnenmacher2022sosp,chen2022linearity,wang2021filter,lin2020hrank}), quantization (\cite{wang2022learnable,liu2022instance,doan2022one}), knowledge distillation (\cite{lin2022knowledge,shu2021channel, binici2022preventing}), neural architecture search (\cite{xiao2022shapley,liu2019darts,zhang2021idarts}), and so on. Among them, neural network pruning has emerged as a promising and effective way to trim and accelerate neural networks significantly without significant or noticeable drops in testing accuracy and/or mean average precision.

Generally, neural network pruning can be categorized as unstructured (\cite{he2022sparse,chen2021long, frankle2019lottery}) and structured (\cite{liu2021group,nonnenmacher2022sosp, wang2021filter,chen2022linearity}). Unstructured pruning masks unwanted individual weights with zeros instead of really removing them. Since these weights still exist in masked neural networks, the memory footprints are not reduced. In addition, masked neural networks often show only a marginal improvement in inference time \cite{wen2016learning,li2017pruning}. Further acceleration requires specific software or hardware \cite{liu2021group, wen2016learning}, but such special support is generally unavailable on resource-constrained devices. In contrast, structured pruning removes entire substructures (such as channels, filters, and layers) and can rebuild a narrower model with a regular structure. Hence it directly speeds up networks and reduces the models’ sizes \cite{liu2021group,nonnenmacher2022sosp,luo2017thinet}. To utilize these advantages, in this paper, we focus on structured pruning from the perspective of channel pruning (equivalent to filter pruning).


\begin{figure*}[ht]
\centering
  \subfloat[Two kinds of loss changes]{
  \begin{minipage}{6cm}
      \includegraphics[width=6cm,height=3.5cm]{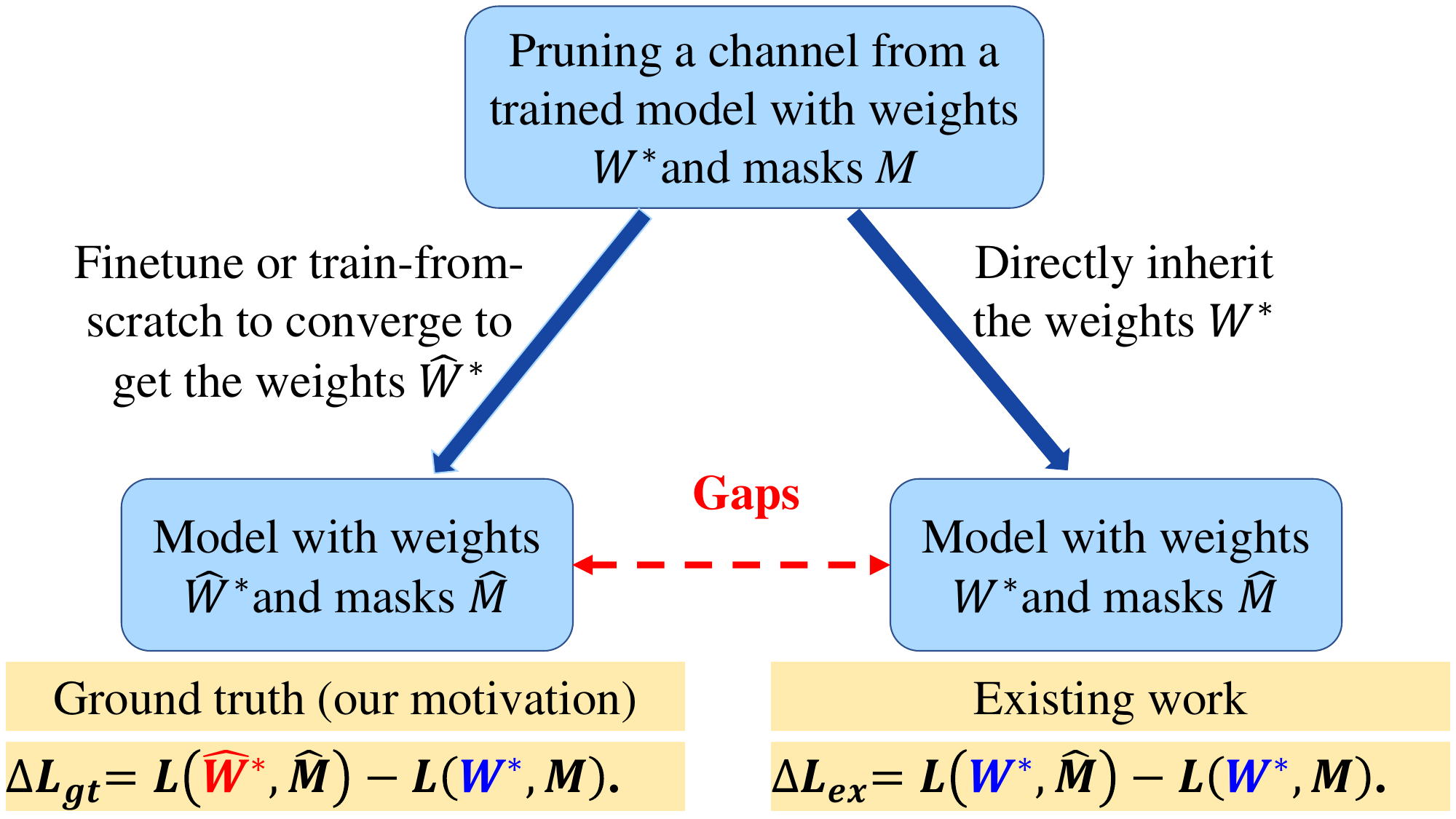}
        \label{fig:differencea}
  \end{minipage}}
  \subfloat[$\Delta L_{gt}$ is different from $\Delta L_{ex}$ significantly]{  
  \begin{minipage}{5.5cm}
       \includegraphics[width=5.5cm,height=3.5cm]{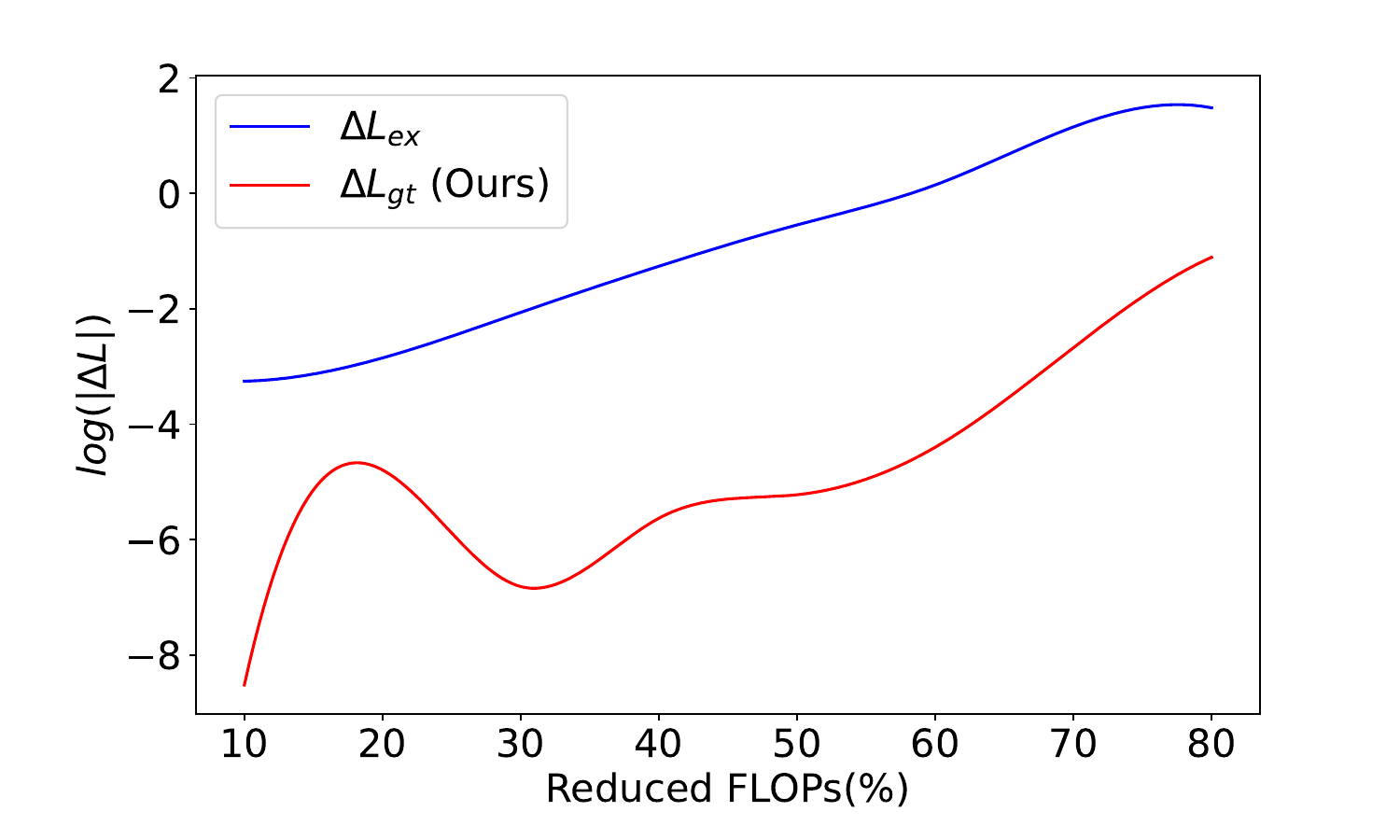}
         \label{fig:differenceb}
  \end{minipage} 
  }
  \subfloat[Non-linear difference of Top-1 accuracy ]{ 
  \begin{minipage}{5.5cm}
       \includegraphics[width=5.5cm,height=3.5cm]{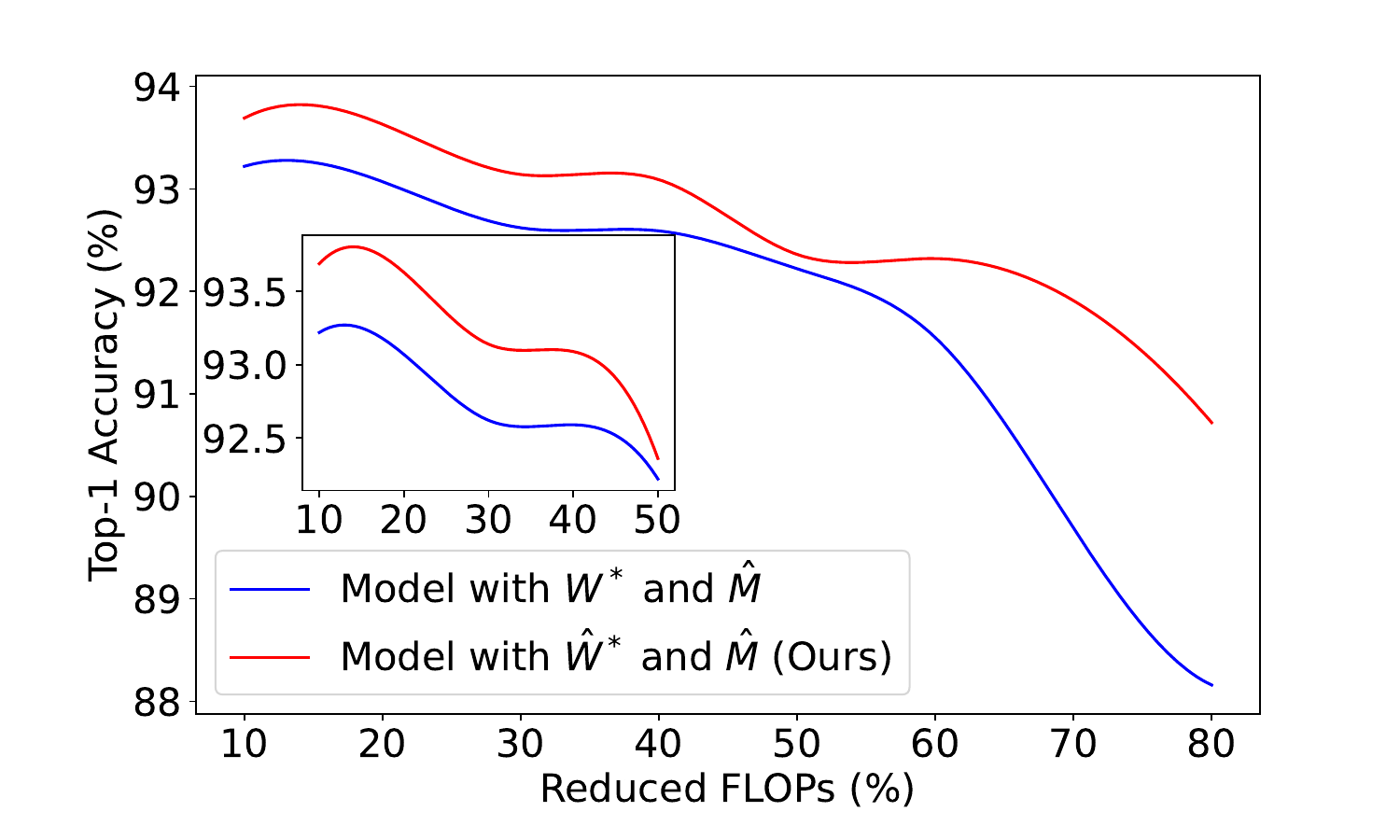}
         \label{fig:differencec}
  \end{minipage} 
  }  
  
  \caption{The difference of loss changes and Top-1 accuracy between with or without retraining of ResNet-32 on CIFAR-10. 
  }
  \label{Fig.1}
  \vspace{-0.3cm}
\end{figure*}

A typical mainstream channel pruning method has three steps: \ding{172} training the dense network to obtain the original model, \ding{173} determining the importance (contribution or redundancy) of channels and pruning channels in the model accordingly, and \ding{174} fine-tuning or training-from-scratch the pruned model to recover its performance. The tasks \ding{173} and \ding{174} are inter-wined. Given model weight variable $W \in \mathbb{R}^{n}$ and pruning mask variable $M \in \{0,1\}^{n}$, where $n$ is the total number of model weights, model pruning can be written as the following bi-level optimization problem with mask $M$ as the upper-level variable and $W$ as the lower-level variable \cite{zhang2022advancing}: 
\begin{equation}
\begin{aligned}
& \mathop{\textrm{min}}\limits_{M \in S} L(W^*(M),M)\\
& \textrm{s.t.}\ W^*(M)=\mathop{\textrm{argmin}}\limits_{W \in \mathbb{R}^{n}} l(M \odot W) + R(W).\\
\end{aligned}
\label{eq:bilevel_optimization}
\end{equation}
For the upper-level optimisation, $L$ denotes the target loss function, $W^*(M)$ denotes the true (updated) optimal weights given mask $M$, and $S=\{M|M \in \{0,1\}^{n}, \textbf{1}^{T}M\leq k\}$ is the feasibility set for $M$ where $k$ is the number of non-zero weights. For the lower-level optimization, $l$ is the training loss (e.g., cross-entropy), and $\odot$ is the element-wise product. $M \odot W$ assigns an element-wise mask $M$ to the model weights $W$. $R$ is the regularization term on $W$. Given a small $k$ that promotes sparsity, the upper-level optimization (corresponding to step \ding{173}) searches for an optimized mask $M$ in the feasibility set of $S$ given model weight variables $W^*$. The discrete property of the mask $M$ makes the problem hard to optimize. A popular paradigm uses various heuristics (e.g., greedy search) to find the least important part to prune. More specifically, in step \ding{173}, for a trained model, each small part of weights (e.g., weights in a single channel) is considered to be separately removed under a new mask $\hat{M}$, and the one with the least loss degradation, 
$\textrm{argmin}_{\hat{M}\subsetneqq M}L(\hat{W}^*, \hat{M})-L(W^*, M)$ is selected.


However, retraining to convergence (corresponding to \ding{174}) after pruning each channel can be prohibitively slow due to the expensive optimization. In existing works (\cite{molchanov2017pruning, you2019gate,He2020learning,liu2021transtailor,nonnenmacher2022sosp}), $\hat{W}^*(\hat{M})$ is simplified as $W^*(\hat{M})$ after the mask is updated from $M$ to $\hat{M}$. Thus, the upper-level optimization problem is changed from 
$\textrm{min}_{\hat{M}\subsetneqq M} L(\hat{W}^*, \hat{M})$ to $\textrm{min}_{\hat{M}\subsetneqq M} L(W^*, \hat{M})$.
For example, evaluating loss changes after pruning each channel is one of the most widely used principles to approximate channel importance. Without loss of generality, we denote the loss change used in the existing methods as $\Delta L_{ex} = L(W^*, \hat{M})-L(W^*, M)$, where $W^*$ is the optimized weights based on channel mask $M$, and $\hat{M}$ is the updated masks from $M$ after masking the pruned channels with zeros. Thus the first loss term of $\Delta L_{ex}$ uses the inherited weights $W^*$ from the trained model with mask $M$ instead of using the optimal weights through retraining based on the updated mask $\hat{M}$ after pruning.

In fact, the ground truth loss change (also called the {\it \textbf{true loss change}} in our paper) should be $\Delta L_{gt} = L(\hat{W}^*, \hat{M})-L(W^*, M)$, where $\hat{W}^*$ is the retrained weights of the pruned model with mask $\hat{M}$. The contrast of the two kinds of loss changes ($\Delta L_{ex}$ and $\Delta L_{gt}$) is pictorially shown in Fig.\ref{Fig.1}(a). Without retraining weights after each pruning, the loss change based on the unchanged weights $W^*$ may \textbf{behave very differently from the true loss change, and produce worse Top-1 accuracy}, especially under a high prune percentage\footnote{Unless otherwise specified, the prune percentage stands for the percentage of the reduced FLOPs compared to the full model in this paper.}, as illustrated in Fig.\ref{Fig.1}(b) and Fig.\ref{Fig.1}(c), respectively. Fig.\ref{Fig.1}(b) shows the apparent difference between $\Delta L_{ex}$ and $\Delta L_{gt}$ through pruning ResNet-32 \cite{he2016deep} on CIFAR-10 \cite{krizhevsky2009learning} as real examples. In Fig.\ref{Fig.1}(b), we remove some specific channels from the fully trained models with varying reduced FLOPs and calculate the loss changes without or with retraining, i.e., $\Delta L_{ex}$ or $\Delta L_{gt}$, respectively. Fig.\ref{Fig.1}(c) shows the corresponding difference of Top-1 accuracy between the model with the unchanged $W^*$ and the model with the retrained $\hat{W}^*$, after removing some channels from the fully trained model. These curves show that the existing loss change calculation is efficient (omitting retraining the models' weights after each pruning) but less unreliable.
To retrain large-scale modern Convolutional Neural Networks (CNNs) to converge after each pruning to get the true loss change would incur unaffordable computations, which is unrealistic in practice. The above observations motivate us to devise a novel, reliable and efficient metric to measure channel importance.

Inspired by \cite{koh2017understanding}, where the influence function is used to measure the influence of each training sample on the model's predictions without retraining, in this paper, we ask the following critical question: {\it \textbf{can we reveal the impact of removing specific channels on the deep models' performance (i.e., the true loss change) without time-consuming retraining?}} If the answer is yes, a series of new paradigms will emerge which differ from the existing pruning methods. To answer this question, we employ influence functions to derive a closed-form solution to estimate the true loss changes without a retraining process for one channel. Furthermore, based on this theoretical derivation, an influence-driven channel sensitivity scoring function 
is derived to evaluate the importance of entire channels \textit{all at once}. Finally, based on the influential scoring function, we propose a novel channel pruning algorithm called {\it Influence Function based Second-Order} ({\textbf{IFSO}}) channel pruning method, which automatically obtains the prune percentage for each layer without the cumbersome yet commonly used sensitivity analysis for local pruning. Extensive experiments demonstrate that the proposed method outperforms existing channel pruning paradigms.

Our contributions can be summarized as follows:
\begin{itemize}
\item \textbf{First}, we observe a non-ignorable gap between the true loss change (with retrained weights) and the existing loss change calculated (with unchanged/old weights) after pruning some specific channels, and this gap degrades the reliability of existing channel importance evaluation, especially under a high prune percentage (see Figure \ref{Fig.1}).
\item \textbf{Second}, to estimate true loss changes avoiding time-consuming retraining, we derive a closed-form solution (see Proposition \ref{prop1}) inspired by influence functions on samples. To the best of our knowledge, this is the first work that utilizes influence functions for deep model pruning.
\item \textbf{Third}, instead of iteratively probing into channel changes, we derive a one-step solution (see Proposition \ref{prop2}) to evaluate the importance of entire channels simultaneously all at once.
\item \textbf{Fourth}, based on our derivations, a novel influence-based channel pruning algorithm, \textbf{\textit{IFSO}} (see Algorithm \ref{alg2}), is developed. Our algorithm globally evaluates channels and automatically obtains the prune percentage per layer.
\item \textbf{Fifth}, we conduct extensive experiments for both image classification and object detection tasks. The experimental results demonstrate that our algorithm outperforms the current state-of-the-arts.
\item \textbf{Sixth}, to the best of our knowledge, our work is the first that shows evaluating true loss changes for pruning without retraining is possible.
This finding will open up opportunities for a series of new paradigms to emerge that differ from existing pruning methods.
\end{itemize}

\section{Related Work}
\label{related_work}
\subsection{Structured Pruning}
Structured pruning is one of the primary ways to compress deep neural networks. For a specific compression constraint, structured pruning aims to remove the least important or redundant substructures (e.g., channels, filters, neurons, or layers) to minimize performance degeneration and maximize acceleration in speed on resource-constrained platforms. Recently, a significant amount of studies have focused on devising efficient metrics to assess the importance of substructures (e.g., channels, filters), such as magnitude (\cite{chen2021long,lee2021layer}), $l_p$ norm (\cite{li2017pruning}), saliency (\cite{zhao2019variational}), distance (\cite{you2020drawing}), geometric medians (\cite{he2019filter}), rank of feature maps (\cite{lin2020hrank}), and reconstruction error (\cite{yu2018nisp,he2017channel,luo2017thinet}). For example, \citet{li2017pruning} calculate filter importance with $l_1$-norm of the filter weights with a widely-used assumption that a weight or feature with a smaller-norm provides less information at the inference time. However, this assumption may not be valid, especially for structured pruning \cite{ye2017rethinking}. Variational CNN Pruning in \cite{zhao2019variational} prunes redundant channels based on distributions of channel saliency measured by the extended scale factors on shift term of the function in the Batch Normalization (BN) layer. Network slimming in \cite{liu2017learning} also reuses BN layer scaling factors where sparsity regularization is imposed during training to automatically identify unimportant channels.
These methods rely on BN and cannot prune sophisticated models such as object detection networks where BN may not be adopted due to the larger input size. \citet{yu2018nisp} propose Neuron Importance Score Propagation (NISP) to minimize the reconstruction error of the final response layer and propagate importance scores through the entire network. However, it still requires the predefined target prune percentage for each layer as the pruning guidance.  


Other sophisticated measures have also been investigated (\cite{He2020learning, liu2021transtailor, dong2017learning,luo2020neural}). Among them, evaluating loss changes (\cite{molchanov2017pruning, you2019gate,He2020learning,liu2021transtailor,nonnenmacher2022sosp}) after pruning substructures is one of the most widely used principles to approximate the substructures' importance. In this line of work, it is generally recognized that less significant substructures have a minor impact on the loss function and hence can be removed. By using Taylor expansion, the loss change induced by pruning can be evaluated approximately. For example, \citet{molchanov2017pruning} propose a first-order pruning method to approximate the loss changes induced by pruning feature maps (or channels) based on grouped activations. \citet{you2019gate} obtain the global filter importance ranking by estimating the loss changes caused by setting the scaling gates of feature maps to zero. In contrast, several second-order pruning methods are proposed to improve performance. For example, \citet{dong2017learning} prune layers based on second-order derivatives of a layer-wise loss function. \citet{nonnenmacher2022sosp} exploit second-order information to select the pruning set of channels to minimize the combined effect on the performance loss of removing all channels in this set. Fisher information also is exploited in \cite{liu2021group,molchanov2019importance} to evaluate the importance of channels via the loss changes induced by discarding them. While first-order methods are more efficient than second-order methods, second-order pruning methods are usually more accurate than first-order methods that neglect possible correlations between different substructures \cite{nonnenmacher2022sosp}. In addition, \citet{He2020learning} develop a differentiable pruning criteria sampler to sample different criteria for different layers by using the criteria loss as a supervision signal. 

Despite their progress, they do not retrain the survived weights after pruning some substructures, thus, prone to unreliable evaluations of channel importance.
As illustrated in Fig.\ref{Fig.1} in Section \ref{sec:introduction}, the gap between the loss changes computed by the existing methods and the true loss changes is hardly neglectable. Calculating the true loss changes after each pruning via retraining the preserved structure to converge each time is too slow. We propose, for the first time, to adopt influence functions to uncover the impact of removing specific channels on the deep model's performance without retraining and derive a closed-form solution to estimate the re-optimized changes in performance loss after pruning each channel efficiently.

\subsection{Influence Functions}
Influence function is a classical technique from robust statistics \cite{cook1980characterizations, hampel1974influence} that reveals how the model parameters change when we upweight or perturb a training sample from the training set $D:=\{z_i=\{x_i,y_i\}\}_{i=1}^{n}$. More specifically, given a model with parameters $W^*$ after trained on $D$, the influence function aims to study the changes of parameters $\Delta W=\hat{W}^*-W^*$ after removing a training sample $z$ and retraining the model parameters on $\hat{D}=D-z$ to $\hat{W}^*$. However, repeatedly retraining the model after removing $z$ is prohibitively time-consuming. 
Therefore, rather than retraining the model after deleting every training sample, \citet{koh2017understanding} estimate $\hat{W}^*$ by minimizing the first-order Taylor series approximation around $W^*$. This way, the influence in the model parameters $W^*$ on up-weighting $z$ can be estimated as:
\begin{equation}
    I(z,W)=\frac{d\hat{W}^*}{d\epsilon}=-H_{W^*}^{-1}\nabla_{W}L(z,W^*),
\label{eq:influence_on_z}    
\end{equation}
where $\epsilon$ is a small scalar and $H_{W^*}:=\frac{1}{n}\sum_{i=1}^{n}\nabla_{W}^{2}L(z_{i},W^*)$ is the Hessian and assumed positive definite. Accordingly, the influence on the loss function of a test sample $z_{test}$ is approximated as:
\begin{equation}
\resizebox{.9\linewidth}{!}{$
I(z,L)=\frac{dL(z_{test},\hat{W}^*)}{d\epsilon}=-\nabla_{W}L(z_{test},W^{*})^{T}H_{W^*}^{-1}\nabla_{W}L(z,W^*).
\label{eq:influence_on_loss}
$}
\end{equation}

The above approximation is similar to leave-one-out retraining \cite{basu2020on}. Thus based on Eq.\eqref{eq:influence_on_z} and Eq.\eqref{eq:influence_on_loss}, we can estimate the validation performance drop after removing a training sample $z$ from $D$ without the expensive process of repeated retraining the model for every removed sample. Since then, there has been increased interest in the applications of influence functions for various machine learning tasks \cite{basu2020on, kong2022resolving}. For example, to understand how model parameters would change when a group of training samples are removed from $D$ and further improve the accuracy of influence functions, \citet{basu2020on} extend influence functions with a second-order approximation to discover the possible cross-correlations among samples and identify influential groups in inference predictions.

Unlike most existing works (\cite{koh2017understanding,basu2020on, kong2022resolving}) which use influence functions to analyze the model parameter changes after removing training samples, this paper adopts influence functions to discover the impact of removing specific channels by estimating true loss changes after each pruning without the time-consuming retraining. To the best of our knowledge, influence functions have not yet been considered for deep model pruning. 



\section{Influence Function Based Second-Order
Channel Pruning by Evaluating True Loss Changes
Without Retraining}
\label{method}
In this section, we will first show how to evaluate the true loss change due to a mask change \textit{without retraining} in a closed form (see Proposition \ref{prop1}) with the estimation quality bounded (see Corollary \ref{corollary1}). 
Iteratively applying this estimator to assess changes from all channels can indeed prune without retraining,
but this iterative process itself is time-consuming and tedious. We thus show how to evaluate the importance of channels all at once (avoiding the iterative process) by further viewing the problem through the lens of sensitivity (see Corollary \ref{corollary2} and Proposition \ref{prop2}), which results in our own sensitivity score (see Definition \ref{definition1}) that captures true loss changes \textit{w.r.t.} both the weights and masks. This leads to our novel global channel pruning algorithm which automatically obtains the prune percentage for each layer. 
\label{method}

\subsection{Evaluating True Loss Changes Without Retraining}
\label{solution1}
According to the bi-level optimization problem defined in Eq.\eqref{eq:bilevel_optimization}, given a model with the optimized weights $W^*$ with mask $M$, it requires retraining the model weights to $\hat{W}^*$ after removing a specific channel for the new mask $\hat{M}$. To bypass the expensive process of repeated retraining,
we use influence functions \cite{basu2020on, koh2017understanding} to estimate how model weights $W^*$ trained with $M$ will change to $\hat{W}^*$ trained with $\hat M$ without retraining.
To achieve this goal, we make a mild and common assumption in our derivation that: \textit{the third and higher derivatives of the loss function \textit{w.r.t.} model weights at optimum are sufficiently small or zero \cite{basu2020on}}. Under this assumption and via the \textbf{second-order} Taylor expansion for $L(\hat{W}^*,\hat{M})$, we can estimate the retrained (truth) loss function changes $\Delta L_{gt}$:
\begin{equation}
\begin{aligned}
    \Delta L_{gt} &= L(\hat{W}^*,\hat{M}) - L(W^*,M) \\
    & = L(W^*,\hat{M}) + \Delta W^{T} \cdot \frac{\partial L(W^*,\hat{M})}{\partial W} \\
    & + \frac{1}{2} \cdot \Delta W^{T} \cdot \frac{\partial^2L(W^*,\hat{M})}{\partial W \partial W} \cdot \Delta W - L(W^*,M),
\end{aligned}
\label{eq:delta_l_gt}
\end{equation}
where $\Delta W = \hat{W}^* - W^*$. Since $\hat{W}^*$ is unknown without time-consuming retraining (e.g., by fine-tuning or training from scratch), we aim to avoid explicitly computing $\Delta W$ through exploiting influence functions \cite{basu2020on,koh2017understanding} to estimate $\hat{W}^*$. 

Based on the implicit function theorem \cite{lorraine2020optimizing}, the retrained true loss function changes $\Delta L_{gt}$ can be estimated in a closed form using the following Proposition.

\begin{prop}
\label{prop1}
Suppose that the deep neural network model obtains the optimized weights $W^*$ with mask $M$ after training it to converge, mask $M$ changes to $\hat{M}$ after removing some specific channels, and the validation loss for $\hat{M}$ is $L(\hat{W}^*,\hat{M})$. If the third and higher derivatives of the loss function \textit{w.r.t.} weights at optimum are zero or sufficiently small, and with $\frac{\partial L(\hat{W^*},\hat{M})}{\partial W} = 0$, we have
\begin{equation}
\resizebox{.85\linewidth}{!}{$
\begin{aligned}
    \Delta L_{gt} &= L(\hat{W}^*,\hat{M}) - L(W^*,M) \\
    &= \Delta L_{ex}-\frac{1}{2} \cdot \frac{\partial L(W^*,\hat{M})}{\partial W}^{T} \cdot \frac{\partial^2L(W^*,\hat{M})^{-1}}{\partial W \partial W} \cdot  \frac{\partial L(W^*,\hat{M})}{\partial W},
\label{eq:prop_1}
\end{aligned}
$}
\end{equation}
where $\Delta L_{ex} = L(W^*, \hat{M})-L(W^*,M)$.
\end{prop}


Proposition \ref{prop1} shows that the true loss change $\Delta L_{gt}$ \textit{w.r.t.} a mask change from $M$ to $\hat{M}$ can be estimated without retraining and without knowing $\hat{W}^*$ explicitly. As shown in Eq.\eqref{eq:prop_1}, $\Delta L_{gt}$ is approximately calculated by combining $\Delta L_{ex}$ and a residual term. Not requiring retraining removes the largest computation hurdle. 
The estimation quality is characterized by Corollary \ref{corollary1} where the approximation error is bounded with two mild and commonly used assumptions in bi-level optimization problems (\cite{couellan2016convergence,zhang2021idarts}). 

\begin{assumption}
\label{assumption1}
$\mathcal{L}(W, M)$ is twice differentiable with constant $C_H$ and is $\lambda$-strongly convex with $W$ around $W^*(M)$. 
\end{assumption}
\begin{assumption}
\label{assumption2}
The $\left \|\frac{\partial L(W^*, \hat{M})}{\partial W}\right \|$ is bounded with constant $C_{l}>0$.
\end{assumption}
\begin{corollary}
\label{corollary1}
Based on the Assumption \ref{assumption1}-\ref{assumption2}, we could bound the error between the approximated validation loss $\mathcal{L}(\hat{W}^*, \hat{M})$ and the ground-truth $\tilde{\mathcal{L}}(\hat{W}^*, \hat{M})$ with $E =\left \|\mathcal{L}(\hat{W}^*, \hat{M}) -\tilde{\mathcal{L}}(\hat{W}^*, \hat{M})  \right \| \leq \frac{\left | \Delta W \right |^3}{6} \textup{max}\left | \frac{\partial \mathcal{L}^3}{\partial W^3} \right |$,
where $\left \| \Delta W \right \| \leq \frac{C_{l}}{\lambda} + \frac{C_H \cdot C_l^2}{2\sigma_{min}^2\lambda}$, $\sigma_{min}$ is the smallest eigenvalue of Hessian matrix $\frac{\partial^2 \mathcal{L}({W}^*, \hat{M})}{\partial W \partial W}$.

\end{corollary}

\subsection{Measure Channel Importance All At Once via Channel Sensitivity}
\label{channelsensitivity}
Proposition \ref{prop1} evaluates the true loss change (without retraining) for a mask change from $M$ to $\hat{M}$. Using it directly or naively for pruning is not yet ideal as it may require repeating such an evaluation \textbf{incrementally} and extensively. 
To see the issue, let's consider a scenario where one wishes to select the 3 least important channels out of $N$ channels in total (where $N>>3$) to prune. In the beginning, the initial mask $M_0$ is the vector of all ones. We can pick a channel to consider (e.g., the first channel) with mask $M_1$ (a vector of zero for that particular channel and ones for the rest). Proposition \ref{prop1} evaluates the true loss change for a mask change from $M_0$ to $M_1$. Next, we can pick another channel to consider (e.g., the second channel) with $M_2$ likewise and evaluate the change from $M_0$ to $M_2$. Once we individually evaluate the changes for all channels, we can sort the changes and pick the channel with the smallest change in the true loss. This means that to select three channels, one may need to evaluate changes for $N+(N-1)+(N-2)$ (not merely 3) times, which can be time-consuming even without retraining.

\textbf{Can we evaluate the importance of entire channels simultaneously all at once, to facilitate one-shot pruning?} The answer is yes. 
To achieve so, we view the importance via the lens of sensitivity (\cite{sun2022disparse,nonnenmacher2022sosp}). Specifically, to evaluate the importance of entire channels at once, we derive solution $\left |\frac{\Delta L_{gt}}{\Delta M}\right |$ to evaluate the influence on the loss function when changing mask $M$, i.e., the sensitivity of $\Delta L_{gt}$ \textit{w.r.t.} $\Delta M=(\hat{M}-M$). 
Following the work in \cite{liu2021group,liu2021transtailor,you2019gate}, we consider the channel mask $M$ to be continuous and make two mild assumptions below, which leads to Corollary \ref{corollary2} that provides a preference of $\Delta M$'s magnitude. 
 \begin{assumption}
\label{assumption3}
For any $W$ and $M$, $\mathcal{L}(\cdot, M)$ and $\mathcal{L}(W, \cdot)$ are Lipschitz continuous with $C_f>0$ and $C_L>0$, respectively.
\end{assumption}
\begin{assumption}
\label{assumption4}
$\left\|\frac{\partial^2 \mathcal{L}({W}^*, M)}{\partial M \partial W}\right\|$ is bounded by $C_{a}>0$. 
\end{assumption}

\begin{corollary}
\label{corollary2}
Based on the Assumption \ref{assumption1}-\ref{assumption4}, and when we assign a continuous perturbation on the mask $M$, we could more tightly bound the error between the approximated validation loss $\mathcal{L}(\hat{W}^*, \hat{M})$ and the ground-truth $\tilde{\mathcal{L}}(\hat{W}^*, \hat{M})$ with $E=\left \|\mathcal{L}(\hat{W}^*, \hat{M}) -\tilde{\mathcal{L}}(\hat{W}^*, \hat{M})  \right \| \leq  \frac{K^3}{6} \textup{max}\left | \frac{\partial \mathcal{L}^3}{\partial W^3} \right |$,
where $K=\frac{C_L}{\lambda}  \left \|\Delta M \right \| + \frac{C_HC_a^2}{2\sigma_{min}^2\lambda} \left \| \Delta M\right \|^2+ o(\left \| \Delta M\right \|^2)$.
\end{corollary}

Corollary \ref{corollary2} shows that 
the approximation error increases with the magnitude of $\Delta M$. Therefore, we pose an infinitesimal change on $M$ to get the new mask $\hat{M}$ when we evaluate the importance of entire channels. 
The following Proposition \ref{prop2} shows the sensitivity of $\Delta L_{gt}$ \textit{w.r.t.} $\Delta M$ after pruning channels without retraining.

\begin{prop}
\label{prop2}
Suppose the channel mask $M$ is continuous, $\Delta M=\hat{M} - M$ is infinitesimally small. Assume that the third and higher order derivatives of the loss function \textit{w.r.t.} $W$ at optimum are sufficiently small. Given a trained network model with the optimized weighs $W^*$ on the mask $M$, the sensitivity of $\Delta L_{gt}$ \textit{w.r.t.} $\Delta M$ can be estimated as:
\begin{equation}
    \resizebox{.85\linewidth}{!}{$\left |\frac{\Delta L_{gt}}{\Delta M}\right |\approx\left |\frac{1}{2} \cdot \Delta M^{T} \cdot \frac{\partial^2 L(W^*,M)}{\partial W \partial M} \cdot H^{-1} \cdot \frac{\partial^2 L(W^*,M)}{\partial M \partial W} \right |,
\label{eq:prop_2}
$}
\end{equation}
where $H=\frac{\partial^2 L(W^*,M)}{\partial W \partial W}$ is the Hessian matrix. 
\end{prop}
More specifically, let $\Delta M = \epsilon \cdot \mathbf{1}$ where $\mathbf{1}$ is a column vector with all ones, and $\epsilon$ is an infinitesimally small scalar, we have
\begin{equation}
    \resizebox{.85\linewidth}{!}{$\left |\frac{\Delta L_{gt}}{\Delta M}\right |\approx\left |\frac{1}{2} \cdot \epsilon \cdot \mathbf{1}^{T} \cdot \frac{\partial^2 L(W^*,M)}{\partial W \partial M} \cdot H^{-1} \cdot \frac{\partial^2 L(W^*,M)}{\partial M \partial W} \right |.
\label{eq:delta_l_respect_m}
$}
\end{equation}
This leads us to define the following sensitivity score vector representing the importance of entire channels all at once.
\begin{definition}
\label{definition1}
(\textbf{Sensitivity Scores}) The sensitivity score vector $S$ \textit{w.r.t.} weights $W$ and mask $M$ is:
\begin{equation}
    S = \left|\mathbf{1}^T \cdot \frac{\partial^2 L(W^*,M)}{\partial W \partial M} \cdot H^{-1} \cdot \frac{\partial^2 L(W^*,M)}{\partial M \partial W}\right|.
\label{eq:sensitivity_score}
\end{equation}
\end{definition}
Let $s_i$ denote the $i$-th entry of $S$ in Eq.\eqref{eq:sensitivity_score}. The value of $s_i$ measures the sensitivity of the loss \textit{w.r.t.} the $i$-th channel's change (regardless of the direction) and reflects the importance of the $i$-th channel. The higher the value, the more significant the channel is. Our sensitivity scores cover all channels and consider second order derivative \textit{w.r.t.} both weights $W$ and mask $M$. The derivative \textit{w.r.t.} weights $W$ reflects the sensitivity to weight changes and is the key to evaluating the true loss change without retraining. The derivative \textit{w.r.t.} mask reflects the sensitivity to mask changes, which is common in other pruning scoring. Next, we will use Group Fisher Scores in \cite{liu2021group} as an example to compare with our proposed scores.

\noindent\textbf{Comparison with Group Fisher Scores} Group Fisher Score $s_i$ of the channel $i$ in \cite{liu2021group} is computed by the sample-wise gradients \textit{w.r.t.} $m_i$: $s_i=\sum_{n=1}^N(\frac{\partial L_n}{\partial m_i})^2$, where $L_n$ denotes the network loss of the $n$-th sample and $m_i$ is an entry in the mask $m$, corresponding to the channel $i$. It only considers the change \textit{w.r.t.} the mask, not the change \textit{w.r.t.} the weights. Not considering weight changes is the main reason existing methods (Group Fisher Scores being an example) can not assess true loss changes, which hinders performance. 

\noindent\textbf{One-shot Pruning vs. Incremental Pruning}
Our proposed sensitivity score $S$ in Eq.\eqref{eq:sensitivity_score} can be utilized for both one-shot and incremental pruning. For an arbitrary prune percentage $p$, one-shot pruning removes channels in a single pass. Specifically, we compute $S$ only once and remove the top $p$ percent of the total channels ranked in ascending order in $S$. The advantage of one-shot pruning is its speed. Its disadvantage is that the sensitivity score for all channels is evaluated at a single $W$. Ideally, after removing one or multiple channels, the optimal $W$ for the rest of the channels can vary and should be updated, resulting in a new $S$. For that reason, one may wish to prune the channels and update $S$ incrementally. 
A practical choice is to prune a batch of channels using an updated $S$, then update $S$ to prune the next batch, and use the accumulated batches to achieve $p$. For example, to prune 1000 channels, one may prune 100 channels each time for a calculated (updated) $S$. Then calculate $S$ and prune 10 times to achieve the target prune percentage. The choice of batch size depends on the size of the network and whether there is a coupling issue of the channels (in some networks, some channels have to be pruned together to maintain integrity due to the internal structure design). For VGGNet \cite{simonyan2015very}, we use batch size 1, which means that to prune 300 channels in VGGNet, we prune one channel each time for a calculated $S$. We then recalculate $S$ and prune the next channel. For ResNet \cite{he2016deep}, due to the coupling issue of some channels, a larger batch size is required. 

The ideas and discussions of one-shot and incremental pruning are not new. It has been observed (as expected) that one-shot pruning has much less computational cost than incremental pruning, but often with some reduction in accuracy \cite{jaiswal2022training,han2015learning}. The batch style of incremental pruning can also be interpreted as a combination of one-shot and incremental pruning, treating processing within a batch as one-shot, and processing batches repeatedly as incremental. We did some ablation study on this in Subsection \ref{combinationofprunings}. 

\subsection{Influence Function based Second-Order Channel Pruning}

Eq.\eqref{eq:sensitivity_score} allows us to evaluate all channels' sensitivity scores simultaneously, based on which we propose a novel channel pruning algorithm named Influence Function based Second-Order (IFSO) presented in Algorithm~\ref{alg2}. The proposed framework can be adapted into both one-shot and incremental pruning paradigms. For one-shot pruning, we prune the specific channels in a single pass to achieve $p$. In contrast, for incremental pruning, we repeat channel pruning actions multiple times. For each pruning action, the number of channels pruned may be one or several, depending on the number of the coupled channels chosen to prune. Channels are removed gradually until the prune percentage $p$ is achieved. It is worth noting that, in pruning complicated structures such as residual connections, the channels in two residual-connected layers are coupled, so they must be discarded or preserved together during the pruning process to avoid breaking the network structures. By default, IFSO in Section \ref{experiments} denotes the incremental pruning method.

As pointed out by \cite{jain2022influence}, computing and storing $H^{-1}$ in Eq.\eqref{eq:sensitivity_score} has $O(h^3)$ and $O(h^2)$ complexity, respectively, where $h$ is the number of parameters in the model (commonly ~100M-100B for a deep model). To improve efficiency, we investigate several commonly used methods of estimating the Inverse-Hessian Vector Products ($H^{-1}v$), including Hessian-free approaches \cite{zhang2022advancing}, Neumann series \cite{lorraine2020optimizing}, and Sherman-Morison approximation \cite{singh2020woodfisher}. In the experiments, we find that the scores computed using Neumann series perform better than those calculated using Sherman-Morison but worse than those computed using the identity matrix (Hessian-free trick) (more discussion can be found in Subsection \ref{approximation_of_inverse_H}), which may seem surprising at first. The Hessian-free trick has been widely used in bi-level optimization applications, e.g., meta-learning \cite{finn2017model} and adversarial learning \cite{zhang2022revisiting}. \citet{jain2022influence} point out that ignoring the Hessian term does not significantly affect rankings by influence. Therefore, we follow this mild and general Hessian-free assumption and replace $H^{-1}$ with the identity matrix in all our experiments. This simplification dramatically improves efficiency and still outperforms competitors.

We accumulate the scores $S$ computed by Eq.\eqref{eq:sensitivity_score} $k$ times for more robust channel scores before executing each pruning action. In our experiments, $k$ is set to 10. More discussion about $k$ can be found in Subsection \ref{pruneinterval}. Besides, to prune less sensitive channels with high memory cost, the raw scores computed by Eq.\eqref{eq:sensitivity_score} are normalized by the memory reduction of pruning channels $\sqrt{\Delta Mem}$. Specifically, we utilize the method in \cite{liu2021group} to compute memory reduction $\Delta Mem$ of pruning channels. 


\begin{algorithm}[t]
\caption{Influence Function based Second-Order(IFSO) Channel Pruning}
\label{alg2}
\hspace*{\algorithmicindent} \textbf{Input:} pre-trained model $W$, training dataset $D$, accumulated times $k$, prune percentage $p$\\
\hspace*{\algorithmicindent} \textbf{Output:} pruned model
\begin{algorithmic}[1]
\STATE \textbf{Initialization:} $M \gets \mathbf{1}$, $S_{accum} \gets 0$
\STATE {find the coupled channels by \cite{liu2021group}}
\WHILE {$p$ is not achieved}
\FOR{$a$ = 1 to $k$}
\STATE {$S_{accum} = S_{accum} + S/\sqrt{\Delta Mem}$, where $S$ is computed by Eq.\eqref{eq:sensitivity_score}}
\STATE {update $W$ once by gradient descent}
\ENDFOR
\STATE {mask the least important channel(s) with zero(s) and $S_{accum} \gets 0$}
\ENDWHILE
\end{algorithmic}
\end{algorithm}

\begin{figure}[t]
\includegraphics[width=8.5cm,height=4.5cm]{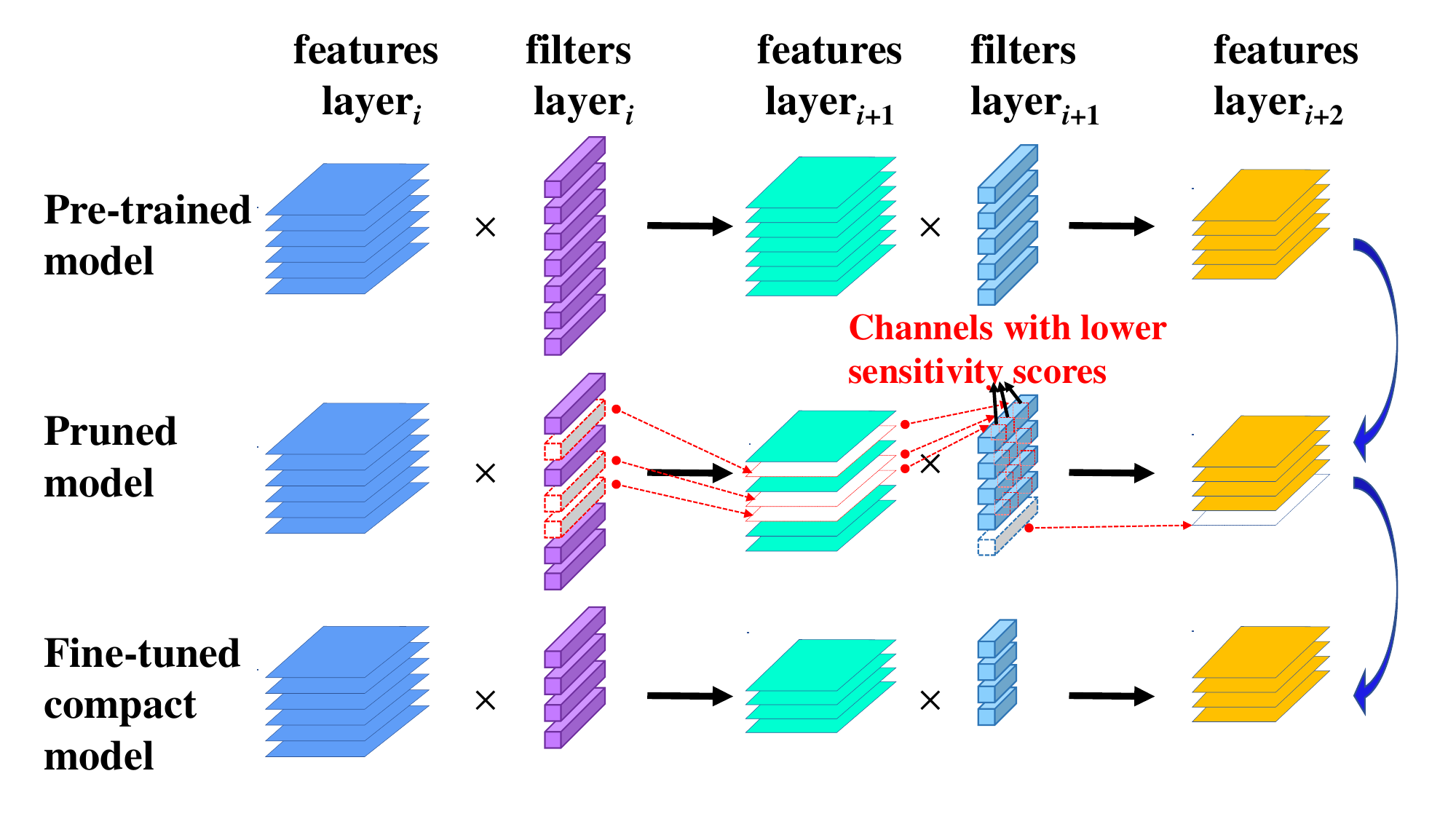}
\caption{Illustration of IFSO channel pruning pipeline (best view in color and zoom in). 
}
\label{Fig.2}
\end{figure}

After the pruning process, we compact and fine-tune the pruned models to recover the performance. The complete pipeline of our proposed IFSO is shown in Fig.\ref{Fig.2}. For the pruning process (as illustrated in the second row), the channels in layer $i+1$ with globally lower sensitivity scores (red dotted rectangle) are chosen and are masked with zero. Then their corresponding filters in layer $i$ are pruned away (red dotted cub), leading to a much smaller model. The red dashed arrows indicate the pruning relationship between filters in layer $i$, input features in layer $i+1$ (i.e., output features in layer $i$), and channels in layer $i+1$. 

\section{Experiments}
\label{experiments}
In this section, we demonstrate the effectiveness of the proposed channel pruning method for image classification and object detection. The section is arranged as follows: in Section \ref{implementation}, we introduce the implementation details, including datasets, network models, the settings for training, pruning and fine-tuning, and evaluation metrics. We present the main results and analysis in Section \ref{resultsandanalyses}. In Section \ref{ablationstudy}, we provide ablation studies for further analysis.
\subsection{Implementation Details}
\label{implementation}
\textbf{Datasets:}
We consider representative datasets and models for image classification and object detection tasks. For image classification, CIFAR-10 \cite{krizhevsky2009learning}, CIFAR-100 \cite{krizhevsky2009learning}, and ImageNet ILSVRC2012 \cite{Russakovsky2015imagenet} are chosen in our experiments. CIFAR-10 and CIFAR-100 datasets contain 50K training images and 10K test images for 10 and 100 classes, respectively. In contrast, ImageNet ILSVRC2012 contains over 1.2 million training images and 50,000 images for validation in 1,000 classes. These two scale datasets differ in their image resolutions (32×32 to 224×224), the number of classes (10 or 100 to 1,000), and the total number of samples (60K to more than 1,000K images). We adopt the large-scale MS COCO 2017 \cite{lin2014microsoft} for object detection. It contains 118K training images and 5K validation images for 80 object categories.

\textbf{Network Models:}
We evaluate our method for image classification task on both single-branch network VGGNet \cite{simonyan2015very} and multiple-branch network ResNet \cite{he2016deep}. Since VGGNet was originally designed for the ImageNet classification tasks, we take a variation of the original VGG-16 for CIFAR-10/CIFAR-100 from \cite{huang2018data}, which consists of 13 convolutional layers and one fully connected (FC) layer. For ResNet on CIFAR-10/CIFAR-100, we choose two different depths, including ResNet-32 and ResNet-56. For ImageNet, we test our algorithm on classic ResNet-50 \cite{he2016deep}. For the object detection task, considering that one-stage detectors have higher inference speeds than two-stage detectors, we prune the classic one-stage model RetinaNet \cite{lin2017focal} to slim and accelerate the model to a higher stage. For all models, we prune channels from all layers except the first convolutional layer.

\textbf{Pre-training, Pruning, and Fine-tuning Settings:}
We adopt SGD as an optimizer with a momentum \cite{sutskever2013on} of 0.9, weight decay of $10^{-4}$. On CIFAR-10/CIFAR-100, we pre-train the models for 200 epochs by using batch size 128 with an initial learning rate of 0.1 and reduce the learning rate at epochs 120 and 160, as used in \cite{nonnenmacher2022sosp}. On ImageNet, we adopt the pre-trained ResNet-50 of MMClassification \cite{mmclassification}. On COCO 2017, we pre-train RetinaNet for 12 epochs by using a batch size of 2 with an initial learning rate of 0.001 and reduce the learning rate after 8 and 11 epochs. 
For pruning all models, we adopt the training datasets as the \textit{proxy datasets} and only use two random batches to calculate channel scores $S$ for efficiency. 
After pruning, we compact and fine-tune the pruned models for image classification and object detection tasks. On ImageNet, we fine-tune the pruned ResNet-50 for 128 epochs by using batch size 256. For the other datasets, we use the same optimization settings as their pre-training phase.

\textbf{Evaluation Metrics:}
We use Top-1 accuracy to evaluate the model performance on CIFAR-10/CIFAR-100. For ImageNet, we report both Top-1 and Top-5 accuracies. In addition, we report the results in mean Average Precision (mAP) for object detection. We adopt the number of FLOPs to evaluate the theoretical acceleration of the pruned models, which is the typical way in the existing works (\cite{liu2021group,dong2017more,he2018soft,he2019filter,liu2017learning,zhao2019variational,wang2021filter,lin2020hrank}). Since operations such as batch normalization (BN) and pooling are insignificant compared to convolution operations, only the number of FLOPs of convolution operations is considered for computation complexity comparison. 

\textbf{Baselines:} The results of the baselines are taken directly from the corresponding papers, which are commonly used in the existing works (\cite{liu2021group,nonnenmacher2022sosp,wang2021filter,lin2020hrank,kang2020operation,He2020learning,you2019gate}). As to GFP \cite{liu2021group} where our IFSO is built upon, we re-implement GFP under the same settings as our method for a fair comparison.

We implement our method on CIFAR-10, CIFAR100, and COCO 2017 by running an NVIDIA GeForce RTX 3090 GPU. For ILSVRC2012, we run ResNet-50 on two NVIDIA GeForce RTX 3090 GPUs. 

\subsection{Main Results and Analysis}
\label{resultsandanalyses}


\subsubsection{Results on CIFAR-10} (1) \textbf{\textit{VGGNet on CIFAR-10}}. We prune VGG-16 on CIFAR-10 with four prune percentages: FLOPs reduced by 40\%, 50\%, 70\%, and 80\%. The comparisons with several state-of-the-art methods are reported in Table~\ref{Prune-VGG16-on-CIFAR10}. The column ``Original'' represents Top-1 accuracy of the unpruned model. ``Pruned'' stands for Top-1 accuracy of the pruned model. ``T1$\downarrow$'' denotes the Top-1 accuracy loss compared with the original model, and smaller is better. ``FLOPs$\downarrow$'' and the suffix numbers in some rows stand for the prune percentages of FLOPs compared to the original model. We report the best Top-1 accuracy of our method. From the results, we see that our method can achieve the best performance under the same or a similar number of FLOPs compared with the previous state-of-the-arts and is still competitive even under a high prune percentage (i.e., 80\% FLOPs reduction). Moreover, our method can achieve equivalent or even better performance than the original deep models under a small or mild prune percentage (e.g., 50-70\% FLOPs reduction). For example, our pruned model outperforms the unpruned baseline by 0.22\% even when the number of FLOPs is reduced by 70\%. In contrast, most comparing baselines encounter performance degradation, where only four competitors can improve Top-1 accuracy under a small prune percentage (i.e., $\leq$50\% FLOPs reduction). For example, CACP achieves the highest 0.51\% Top-1 accuracy improvement among the four competitors when only 30\% FLOPs is reduced, but our method can achieve the highest 0.70\% Top-1 accuracy improvement with 40\% FLOPs reduction.

\begin{table}[t]
\centering
  \caption{Performance of pruning VGG-16 on CIFAR-10 under different prune percentages.}
  \label{Prune-VGG16-on-CIFAR10}
   \begin{tabular}{l|c|ccc} 
    \Xhline{0.3ex}
    \multirow{2}{*}{Method} & FLOPs $\downarrow$ & Original & Pruned & T1 $\downarrow$ \\
    & (\%) & T1(\%) & T1(\%) & (\%) \\
    \hline
    Pruned-A \cite{wang2021filter} & 11.6 & 93.03 & 93.18 & -0.15 \\
    CACP-30 \cite{liu2021conditional} & 30.0 & 93.02 & 93.53 & \underline{-0.51} \\
    Chen \cite{chen2019shallowing} & 38.9 & 93.50 & 93.40 & 0.10 \\
    VCP \cite{zhao2019variational} & 39.1 & 93.25 & 93.18 & 0.07 \\
    SSS \cite{huang2018data} & 41.6 & 93.96 & 93.02 & 0.94 \\
    GFP-40 \cite{liu2021group} & 40.0 & 93.34 & \underline{93.72} & -0.38 \\    
    IFSO-40 (\textbf{ours}) & 40.0 & 93.34 & \textbf{94.04} & \textbf{-0.70} \\ 
    \hline
    Slimming \cite{liu2017learning} & 48.1 & 93.85 & 92.91 & 0.94 \\
    CC \cite{li2021towards} & 50.8 & 93.70 & \textbf{94.15} & -0.45 \\
    GFP-50 \cite{liu2021group} & 50.0 & 93.34 & 93.81 & \underline{-0.47} \\    
    IFSO-50 (\textbf{ours}) & 50.0 & 93.34 & \underline{93.94} & \textbf{-0.60} \\
    \hline
    SCP \cite{kang2020operation} & 66.2 & 93.85 & \textbf{93.79} & 0.06 \\
    CACP-70 \cite{liu2021conditional} & 70.0 & 93.02 & 92.89 & 0.13 \\
    GFP-70 \cite{liu2021group} & 70.0 & 93.34 & 93.48 & \underline{-0.14} \\
    IFSO-70 (\textbf{ours}) & 70.0 & 93.34 & \underline{93.56} & \textbf{-0.22} \\
    \hline
    GFP-80 \cite{liu2021group} & 80.0 & 93.34 & 93.03 & 0.31 \\
    IFSO-80 (\textbf{ours}) & 80.0 & 93.34 & \textbf{93.31} & \textbf{0.03} \\
    \Xhline{0.3ex}
  \end{tabular}
\flushleft{The best result is shown in \textbf{bold} and the second best is \underline{underlined}.}
  \vspace{-0.2cm}
\end{table}

(2) \textbf{\textit{ResNet on CIFAR-10}}. 
Unlike VGGNet, ResNet is more compact and has less redundancy. Therefore, channel pruning for ResNet seems to be more challenging. For CIFAR-10, we test our method on ResNet-32 and ResNet-56 with two prune percentages: FLOPs reduced by 40\% and 50\%. For ResNet-32, from Table~\ref{Prune-Resnet32-on-CIFAR10}, our method achieves the lowest Top-1 accuracy drops under the equivalent or a similar number of FLOPs. For example, when the number of FLOPs is reduced by 50.1\%, our method has the lowest reduction of 0.51\% in Top-1 accuracy. Besides, compared with MIL \cite{dong2017more} and GFP-40 \cite{liu2021group}, our method can obtain a lower Top1 accuracy drop with a higher speedup. 

\begin{table}[t]
  \centering
  \caption{Performance of pruning ResNet-32 on CIFAR-10 under different prune percentages.}
  \label{Prune-Resnet32-on-CIFAR10}
   \begin{tabular}{l|c|ccc} 
    \Xhline{0.3ex}
    \multirow{2}{*}{Method} & FLOPs $\downarrow$ & Original & Pruned& T1 $\downarrow$ \\
    & (\%) & T1(\%) & T1(\%) & (\%) \\ 
    \hline
    MIL \cite{dong2017more} & 31.2 & 92.33 & 90.74 & 1.59 \\
    SFP \cite{he2018soft} & 41.5 & 92.63 & 92.08 & 0.55 \\
    FPGM \cite{he2019filter} & 41.5 & 92.63 & 92.31 & \underline{0.32} \\
    GFP-40 \cite{liu2021group} & 40.1 & 93.51 & \underline{92.86} & 0.65 \\ 
    IFSO-40 (\textbf{ours}) & 40.2 & 93.51 & \textbf{93.23} & \textbf{0.28} \\
    \hline
    PScratch \cite{wang2020pruning} & 50.0 & 93.18 & 92.18 & 1.00 \\
    GAL \cite{lin2019towards} & 50.0 & 93.18 & 91.72 & 1.46 \\
    GFP-50 \cite{liu2021group} & 50.1 & 93.51 & \underline{92.64} & \underline{0.87} \\
    IFSO-50 (\textbf{ours}) & 50.1 & 93.51 & \textbf{93.00} & \textbf{0.51} \\
    \Xhline{0.3ex}
  \end{tabular}
  \flushleft{Unless otherwise specified, the columns, the suffix number in some rows, bold and underlined of this table and other tables have the same meaning as Table~\ref{Prune-VGG16-on-CIFAR10}.}
  \vspace{-0.2cm}
\end{table}

\begin{figure}[t]
\centering
\begin{minipage}{9.5cm}
\includegraphics[width=9.5cm,height=5cm]{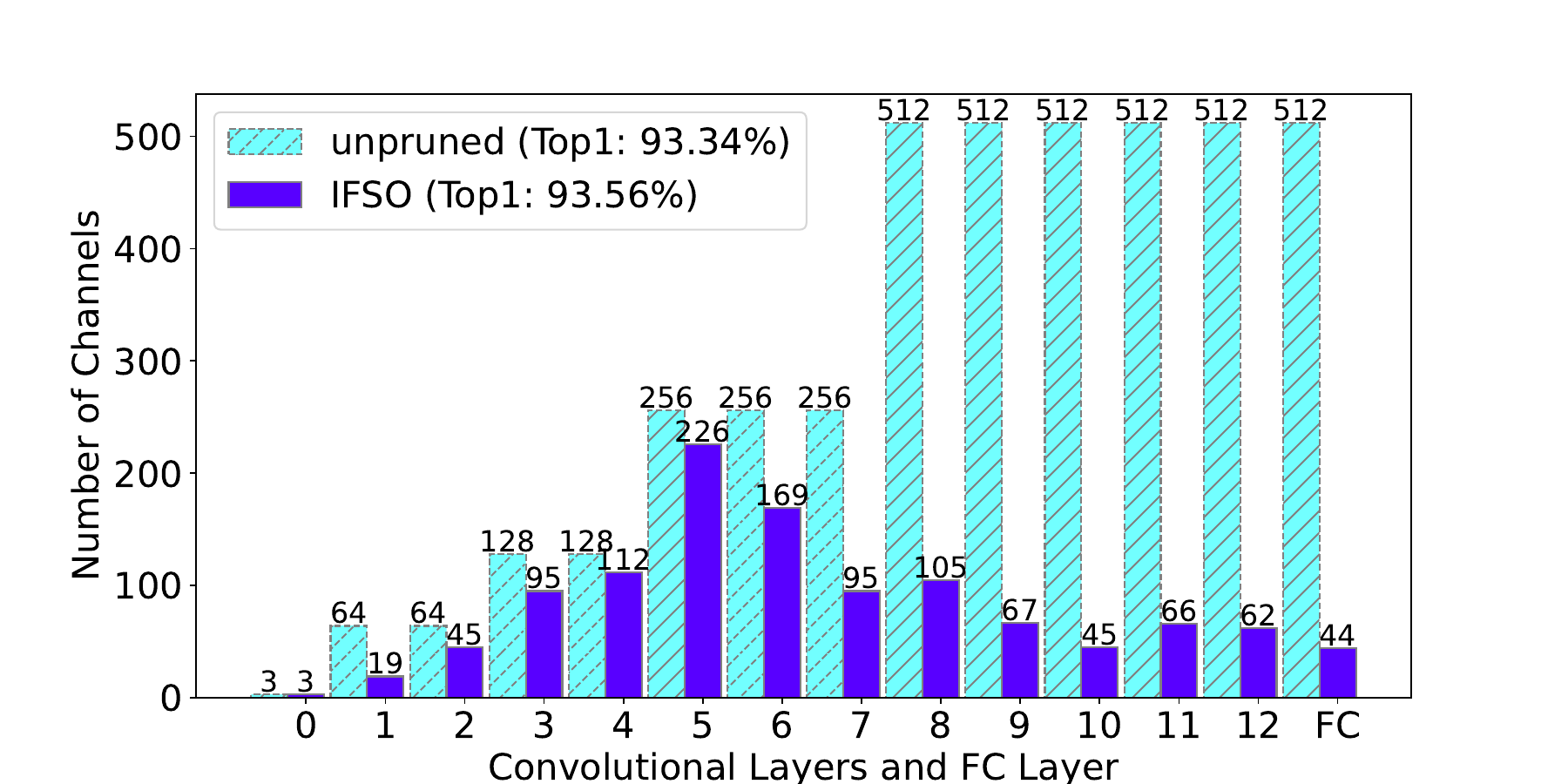}
\end{minipage}
\caption{Visualization of the pruned VGG-16 on CIFAR-10 with 70\% FLOPs reduced. }
\label{Fig.3}
\end{figure}

For ResNet-56 on CIFAR-10, our method has the lowest performance degradation than the existing state-of-the-art methods under the equivalent or a similar percentage of FLOPs reduction as shown in Table~\ref{Prune-ResNet56-on-CIFAR10}. For example, our pruned model outperforms the unpruned baseline by 0.27\% when the number of FLOPs is reduced by 40.2\%. On the other hand, although HRank has a very close result of -0.26\%, the number of FLOPs is only reduced by 29.3\%.

\begin{table}[t]
\centering
\caption{Performance of pruning ResNet-56 on CIFAR-10 under different prune percentages.}
\label{Prune-ResNet56-on-CIFAR10}
\begin{tabular}{l|c|ccc}
\Xhline{0.3ex}
\multirow{2}{*}{Method} & FLOPs $\downarrow$ & Original & Pruned& T1 $\downarrow$ \\
& (\%) & T1(\%) & T1(\%) & (\%) \\ 
\hline
Li \textit{et al.} \cite{li2017pruning} & 27.6 & 93.04 & 93.06 & -0.02 \\
HRank \cite{lin2020hrank} & 29.3 & 93.26 & 93.52 & \underline{-0.26} \\
Chen \cite{chen2019shallowing} & 34.8 & 93.03 & 93.09 & -0.06 \\
NISP \cite{yu2018nisp} & 35.5 & 93.26 & 93.01 & 0.25 \\
GAL \cite{lin2019towards} & 37.6 & 93.26 & 93.38 & -0.12 \\
GFP-40 \cite{liu2021group} & 40.2 & 93.68 & \underline{93.54} & 0.14 \\
IFSO-40 (\textbf{ours}) & 40.2 & 93.68 & \textbf{93.95} & \textbf{-0.27} \\
\hline
He \textit{et al.} \cite{he2017channel} & 50.0 & 93.26 & 90.80 & 2.46 \\
AMC \cite {he2018amc} & 50.0 & 92.80 & 91.90 & 0.90 \\
DCP \cite{zhuang2018discrimination} & 50.0 & 93.80 & 93.49 & 0.31 \\
FPGM \cite{he2019filter} & 52.6 & 93.59 & 93.26 & 0.33 \\
LFPC \cite{He2020learning} & 52.9 & 93.59 & 93.24 & 0.35 \\
GFP-50 \cite{liu2021group} & 50.0 & 93.68 & \underline{93.53} & \underline{0.15} \\
IFSO-50 (\textbf{ours}) & 50.0 & 93.68 & \textbf{93.65} & \textbf{0.03} \\
\Xhline{0.3ex}
\end{tabular}
\end{table}

\begin{table}[t]
\centering
  \caption{Performance of pruning VGG-16 on CIFAR-100 under different prune percentages.}
  \label{Prune-VGG16-on-CIFAR100}
  \centering
  \begin{tabular}{l|c|ccc} 
  \Xhline{0.3ex}
  \multirow{2}{*}{Method} & FLOPs $\downarrow$ & Original & Pruned& T1 $\downarrow$ \\
  & (\%) & T1(\%) & T1(\%) & (\%) \\
  \hline
    GBN-40 \cite{you2019gate} & 40.0 & 73.20 & \textbf{73.00} & \underline{0.20} \\
    GFP-40 \cite{liu2021group} & 40.0 & 72.88 & 72.49 & 0.39 \\
    IFSO-40 (\textbf{ours}) & 40.0 & 72.88 & \underline{72.93} & \textbf{-0.05} \\
    \hline
    GBN-50 \cite{you2019gate} & 50.0 & 73.20 & 71.40 & 1.80 \\
    GFP-50 \cite{liu2021group} & 50.0 & 72.88 & \underline{72.30} & \underline{0.58} \\    
    IFSO-50 (\textbf{ours}) & 50.0 & 72.88 & \textbf{72.83} & \textbf{0.05} \\
    \Xhline{0.3ex}
  \end{tabular}
\end{table}

\begin{table*}[t]
  \caption{Performance of pruning ResNet-50 on ILSVRC2012 under different prune percentages.}
  \label{Prune-Resnet50-on-ILSVRC2012}
  \centering
   \begin{tabular}{l|c|ccc|ccc} 
    \Xhline{0.3ex}
    \multirow{2}{*}{Method} & FLOPs $\downarrow$ & Original & Pruned & T1 $\downarrow$ & Original & Pruned & T5 $\downarrow$ \\
    & (\%) & T1(\%) & T1(\%) & (\%) & T5(\%) & T5(\%) & (\%) \\
    \hline
    SSS-31 \cite{huang2018data} & 31.1 & 76.12 & 74.18 & 1.94 & 92.86 & 91.91 & 0.95 \\
    ThiNet \cite{luo2017thinet} & 36.8 & 75.30 & 74.03 & \underline{1.27} & 92.20 & \underline{92.11} & \textbf{0.09} \\
    SFP \cite{he2018soft} & 41.8 & 76.15 & \underline{74.61} & 1.54 & 92.87 & 92.06 & 0.81 \\ 
    IFSO-40 (\textbf{ours}) & 40.0 & 76.55 & \textbf{75.70} & \textbf{0.85} & 93.06 & \textbf{92.74} & \underline{0.32} \\
    \hline
    SSS-43 \cite{huang2018data} & 43.0 & 76.12 & 71.82 & 4.30 & 92.86 & 90.79 & 2.07 \\
    Taylor-FO-BN \cite{molchanov2019importance} & 45.0 & 76.18 & \underline{74.50} & \underline{1.68} & - & - & - \\ 
    GDP-45 \cite{lin2018accelerating} & 45.2 & 75.13 & 72.61 & 2.52 & 92.30 & \underline{91.05} & \underline{1.25} \\ 
    GDP-51 \cite{lin2018accelerating} & 51.3 & 75.13 & 71.89 & 3.24 & 92.30 & 90.71 & 1.59 \\ 
    IFSO-50 (\textbf{ours}) & 50.0 & 76.55 & \textbf{75.08} & \textbf{1.47} & 93.06 & \textbf{92.28} & \textbf{0.78} \\
    \Xhline{0.3ex}
  \end{tabular}
  \flushleft{``T5'' represents Top-5 accuracy. The meaning of other columns and number suffixes refer to Table~\ref{Prune-VGG16-on-CIFAR10}. ``-'' means the corresponding result is not reported.}
\end{table*}

\begin{table}[t]
  \caption{Performance of pruning ResNet-56 on CIFAR-100 under different prune percentages.}
  \label{Prune-ResNet56-on-CIFAR100}
  \centering
  \begin{tabular}{l|c|ccc} 
  \Xhline{0.3ex}
  \multirow{2}{*}{Method} & FLOPs $\downarrow$ & Original & Pruned& T1 $\downarrow$ \\
  & (\%) & T1(\%) & T1(\%) & (\%) \\
  \hline
    MIL \cite{dong2017more} & 39.3 & 71.33 & 68.37 & 2.96 \\
    GFP-40 \cite{liu2021group} & 40.0 & 71.95 & \underline{71.37} & \underline{0.58} \\
    IFSO-40 (\textbf{ours}) & 40.0 & 71.95 & \textbf{71.56} & \textbf{0.39} \\
    \hline
    SFP \cite{he2018soft} & 52.6 & 71.40 & 68.79 & 2.61 \\
    FPGM \cite{he2019filter} & 52.6 & 71.41 & 69.66 & \underline{1.75} \\
    GFP-50 \cite{liu2021group} & 50.0 & 71.95 & \underline{70.17} & 1.78 \\     
    IFSO-50 (\textbf{ours}) & 50.0 & 71.95 & \textbf{70.55} & \textbf{1.40} \\
    \Xhline{0.3ex}
  \end{tabular}
\end{table}

\subsubsection{Results on CIFAR-100}
For CIFAR-100, we test our method on VGG-16 and ResNet-56 with two prune percentages: FLOPs reduced by 40\% and 50 \%. The results of pruning VGG-16 and ResNet-56 on CIFAR-100 are shown in Table~\ref{Prune-VGG16-on-CIFAR100} and Table~\ref{Prune-ResNet56-on-CIFAR100}, respectively. From Table~\ref{Prune-VGG16-on-CIFAR100}, at the equivalent FLOPs threshold, the Top-1 accuracy loss of our method is lower than those pruned by GBN \cite{you2019gate} and GFP \cite{liu2021group}. For example, when reducing 40\% FLOPs, our pruned model achieves 72.93\% Top-1 accuracy, which is even higher than the unpruned model, while GBN \cite{you2019gate} has a reduction of 0.20\% in Top-1 accuracy. From the experimental results shown in Table~\ref{Prune-ResNet56-on-CIFAR100}, the performance degradation of our pruned ResNet56 is also smaller than the competitors. For example, the Top-1 accuracy loss of our pruned model is 2.57\% lower than that of MIL \cite{dong2017more} even with higher FLOPs reduction.

\subsubsection{Results on ILSVRC2012}
To validate the effectiveness of the proposed method on large-scale datasets, we further perform our method on the widely used ResNet-50 \cite{he2016deep} on ImageNet ILSVRC2012 \cite{Russakovsky2015imagenet} with two prune percentages: FLOPs reduced by 40\% and 50\%. Compared with other methods under a similar number of FLOPs, as shown in Table~\ref{Prune-Resnet50-on-ILSVRC2012}, our method achieves the lowest reduction of 0.85\% and 1.47\% in Top-1 accuracy under a similar number of FLOPs, respectively. Furthermore, although ThiNet \cite{luo2017thinet} has a lower drop in Top-5 accuracy when 36.8\% FLOPs is reduced than our method when 40\% FLOPs is dropped, our method achieves 0.42\% lower drop in Top-1 accuracy as well as a higher FLOPs reduction percentage. 

\subsubsection{Results on COCO} 
Due to the larger input size and more complicated network architectures, pruning for object detection is more challenging than the task for image classification. As a result, most existing channel pruning works (\cite{nonnenmacher2022sosp,he2019filter,li2021towards,kang2020operation,luo2020neural,lin2020hrank,He2020learning}) only report their results for image classification. To further demonstrate the effectiveness of our method, we conduct object detection experiments on COCO 2017 based on the detection framework MMDetection \cite{chen2018mmdetection}. We prune RetinaNet with two prune percentages: FLOPs reduced by 50\% and 60\%. 

\begin{table*}[t]
  \caption{Performance of pruning RetinaNet on COCO 2017 under different prune percentages.}
  \label{Prune-RetinaNet-on-COCO}
  \centering
  \begin{tabular}{l|cc|cccccccc}
    \Xhline{0.3ex}
    \multirow{2}{*}{Method} & FLOPs$\downarrow$ & GPU & Original & Pruned & mAP$\downarrow$ & mAP50 & mAP75 \\
    & (\%) & (GB) & (\%) & (\%) & (\%) & (\%) & (\%) \\
    \hline
    GFP-50 \cite{liu2021group} & 50.0 & 6.5 & 36.10 & 36.20 & -0.10 & 55.50 & 38.50 \\
    IFSO-50 (\textbf{ours}) & 50.0 & \textbf{6.0} & 36.10 & \textbf{36.30} & \textbf{-0.20} & \textbf{55.60} & \textbf{38.70} \\
    \hline
    GFP-60 \cite{liu2021group} & 60.0 & 6.1 & 36.10 & 35.30 & 0.80 & 54.60 & 37.40 \\
    IFSO-60 (\textbf{ours}) & 60.0 & \textbf{5.6} & 36.10 & \textbf{35.70} & \textbf{0.40} & \textbf{54.70} & \textbf{38.00} \\
    \Xhline{0.3ex}
  \end{tabular}
  \flushleft{The column ``Original'' and ``Pruned'' are measured in mAP. ``mAP$\downarrow$'' denotes the mAP loss compared with the original model, and smaller is better. ``GPU'' denotes the GPU consumption of the pruned model during the fine-tuning stage, and smaller is better. The meaning of ``FLOPs$\downarrow$'' and number suffixes refer to Table~\ref{Prune-VGG16-on-CIFAR10}.}
\end{table*}

\begin{figure}[t]
\centering
\begin{minipage}{8.5cm}
\includegraphics[width=8.5cm,height=4.5cm]{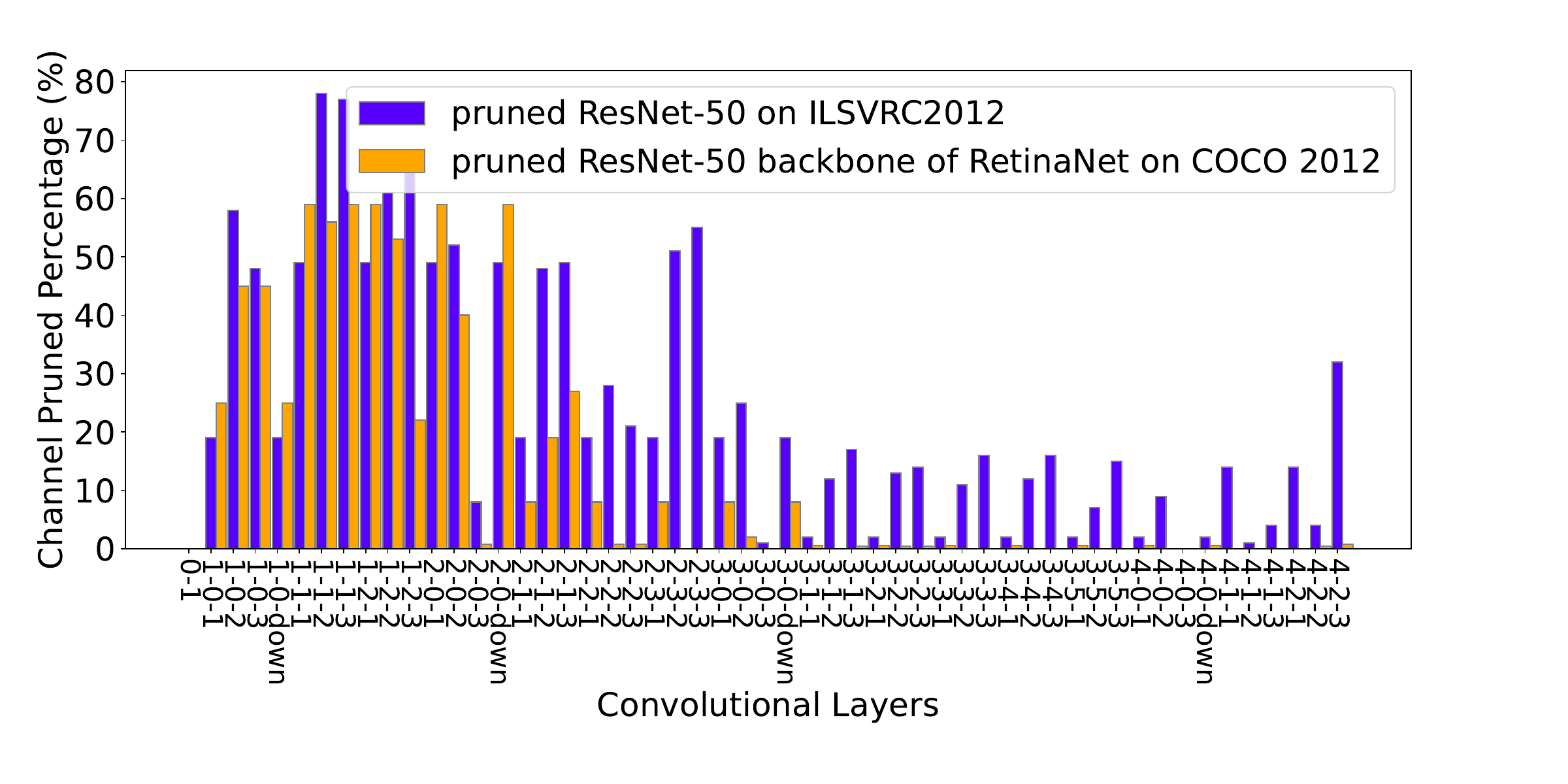}
\end{minipage}
\caption{An illustration of the pruned ResNet50. (This figure is best viewed in color and zoomed in.)}
\label{Fig.4}
\vspace{-0.3cm}
\end{figure}

As shown in Table~\ref{Prune-RetinaNet-on-COCO}, our pruned model outperforms the unpruned baseline by 0.20\% mAP even when the number of FLOPs is reduced by 50\%. Although our Top-1 accuracy is slightly higher than that of GFP \cite{liu2021group} when 50\% FLOPs is dropped, our pruned model consumes less GPU memory than those of GFP's pruned model. 
In addition, our method achieves better performance than GFP \cite{liu2021group} when 60\% FLOPs is reduced.

\subsubsection{Pruned Structure Visualization} 
Fig.~\ref{Fig.3} shows the detailed structure of a pruned VGG-16 obtained from our method on CIFAR-10 with 70\% FLOPs reduced. Compared with the original model, the pruned VGG-16 obviously has lower complexities, while the pruned model outperforms the original model by 0.22\% in Top-1 accuracy. In addition, we find that most of the channels in the deep layers are insignificant, while the channels in the middle layers seem more critical.

In our experiments, ResNet-50 is used for both image classification on ILSVRC2012 and the backbone of object detector, i.e., RetinaNet, on COCO 2017. To explore the pruning structure of the same model applied to different tasks, in Fig.~\ref{Fig.4}, we visualize the structures of the pruned ResNet-50 on ILSVRC2012 and COCO 2017 when 50\% FLOPs is reduced, respectively. The index ``a-b'' or ``a-b-c'' indicates the residual block where the convolutional layer is located. In contrast to VGGNet, the pruned ResNet-50 for both tasks keep more channels in deep layers. In addition, for detection, the pruned ResNet-50 backbone keeps the whole channels of many convolutions in deep layers, as detection needs to keep more feature information for the later stages.

We also show the detailed structure of the pruned RetinaNet on COCO 2017 in Table~\ref{detailed-RetinaNet-on-COCO-2017-dataset}. Since the detailed structure of the pruned backbone of RetinaNet is shown in Fig.~\ref{Fig.4}, here we only show the rest components of RetinaNet. We can see that most of the channels in the head are insignificant, while the channels in the neck seem more critical.

\begin{table}[t]
  \caption{Detailed structure of the pruned RetinaNet (except backbone) on COCO 2017 with 50\% FLOPs reduced.}
  \label{detailed-RetinaNet-on-COCO-2017-dataset}
  \centering
  \begin{tabular}{l|ccc}
  \Xhline{0.3ex}
    \multirow{2}{*}{Layer} & \#Channels & \#Remained & Pruning rate \\
    & & Channels & (\%) \\
    \hline
    neck lateral-0 & 512 & 471 & 8.00\\
    neck lateral-1 & 1024 & 1018 & 0.59\\
    neck lateral-2 & 2048 & 2048 & 0\\ 
    neck fpn-0 & 256 & 255 & 0.39\\
    neck fpn-1 & 256 & 255 & 0.39\\
    neck fpn-2 & 256 & 255 & 0.39\\
    neck fpn-3 & 256 & 256 & 0\\
    neck fpn-4 & 256 & 256 & 0\\
    head cls-0 & 256 & 256 & 0\\
    head cls-1 & 256 & 136 & 46.88\\
    head cls-2 & 256 & 107 & 58.20\\
    head cls-3 & 256 & 99 & 61.33\\ 
    head reg-0 & 256 & 256 & 0\\ 
    head reg-1 & 256 & 130 & 49.22\\ 
    head reg-2 & 256 & 93 & 63.67\\ 
    head reg-3 & 256 & 86 & 66.41\\
    head retina-cls & 256 & 77 & 69.92\\
    head retina-reg & 256 & 89 & 65.23\\
    \Xhline{0.3ex}
  \end{tabular}
\end{table}

\subsubsection{Performance with Different prune percentages} 
\begin{figure*}[th]
\centering
  \subfloat[Prune ResNet-32 on CIFAR-10]{
  \begin{minipage}{4.7cm}
      \includegraphics[width=4.8cm,height=3.6cm]{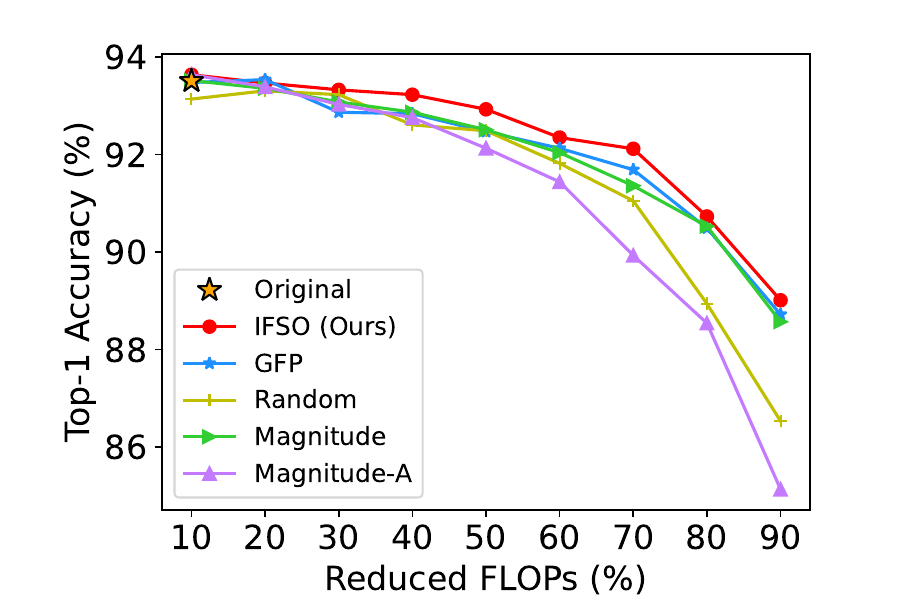}
        \label{fig5a:ResNet-32}
  \end{minipage}} \hspace{-5mm}
  \subfloat[Prune VGG-16 on CIFAR-10]{
  \begin{minipage}{4.7cm}
       \includegraphics[width=4.8cm,height=3.6cm]{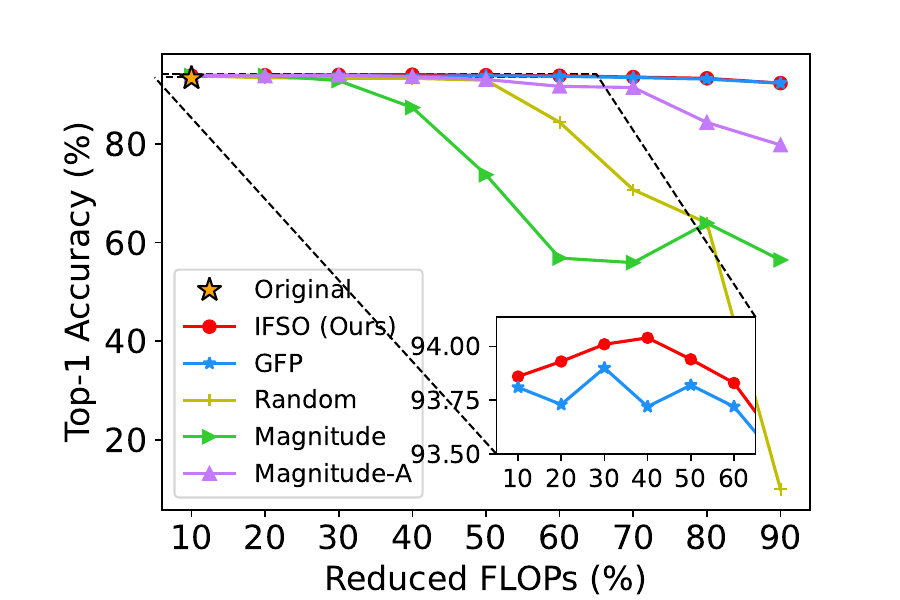}
         \label{fig5b:VGG-16}
  \end{minipage} 
  } \hspace{-6mm}
  \subfloat[Prune ResNet-32 on CIFAR-100]{ 
  \begin{minipage}{4.7cm}
       \includegraphics[width=4.8cm,height=3.6cm]{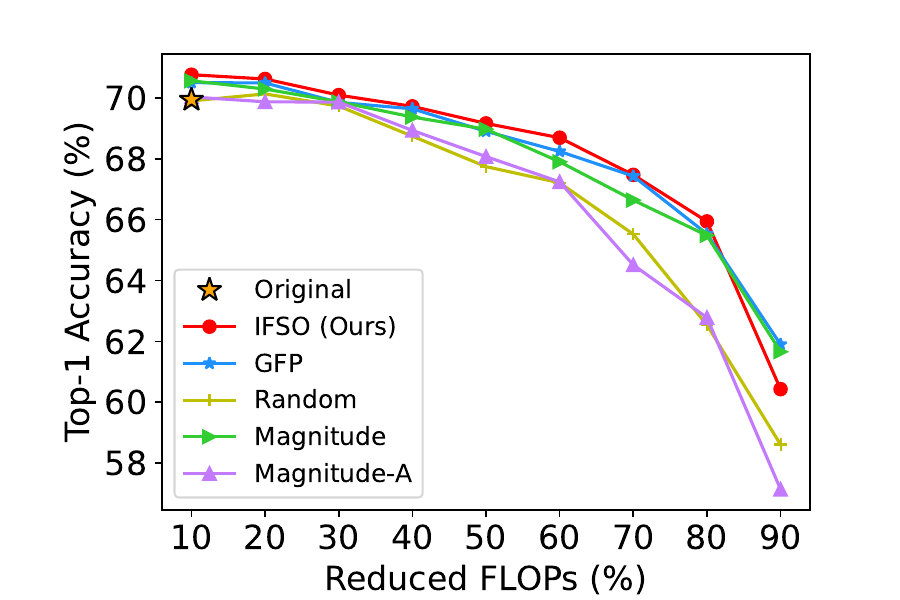}
         \label{fig5c:ResNet-32}
  \end{minipage} 
  } \hspace{-6mm}
  \subfloat[Prune VGG-16 on CIFAR-100]{ 
  \begin{minipage}{4.7cm}
       \includegraphics[width=4.8cm,height=3.6cm]{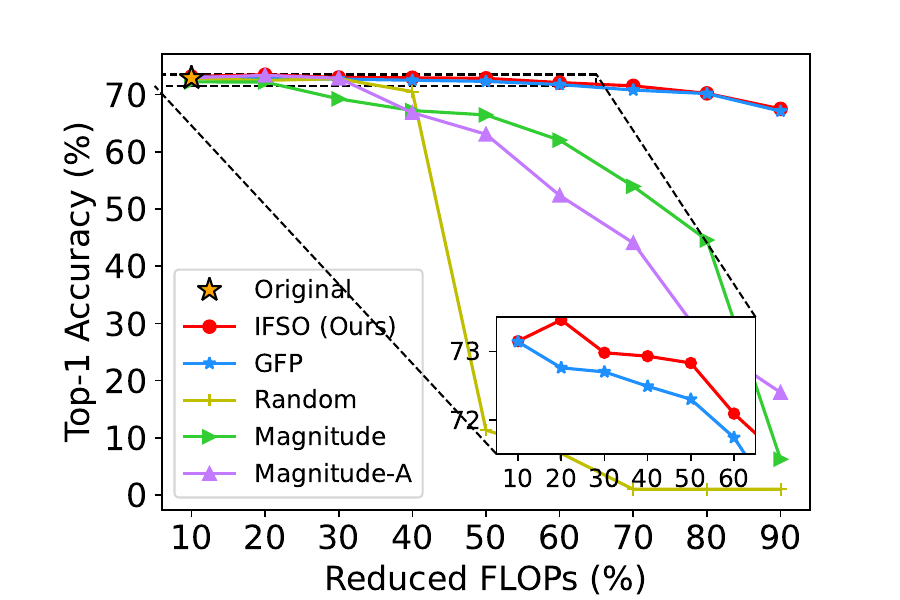}
         \label{fig5d:VGG-16}
  \end{minipage} 
  } \\ 
 \caption{Performance comparison of different channel selection methods with pruned FLOPs from 10\% to 90\%. Our method performs better except in rare cases. (This figure is best viewed in color and zoomed in.)}
 \label{Fig.5}
\vspace{-0.3cm}
\end{figure*}

We summarize the performance of our IFSO across different prune percentages in Fig.~\ref{Fig.5}, which displays Top-1 accuracy of different channel pruning methods versus the reduced FLOPs percentage for ResNet-32/VGG-16 on CIFAR-10/CIFAR-100. ``Original'' refers to the unpruned model. ``Random'' refers to that channels are randomly removed. ``Magnitude'' calculates the sum of the weights in each channel $i$ as its importance score: $s_i=\sum\left| W(:,i,:,:)\right|$. The suffix ``A'' in ``Magnitude-A'' stands for ``Average''. ``Magnitude-A'' calculates the scores as: $s_{ki}=\frac{1}{N_k}\sum\left| W(:,i,:,:)\right|$, where $N_{k}$ is the number of filters in layer $k$. As shown in Fig.~\ref{Fig.5}, IFSO achieves higher Top-1 accuracy compared with other methods. In some cases, models pruned with IFSO even outperform unpruned models. For example, our pruned model outperforms the unpruned baseline even when the number of FLOPs is reduced by 70\% on VGG-16 on CIFAR-10. Both magnitude methods generally have lower Top-1 accuracy than GFP \cite{liu2021group} and IFSO. In addition, we observe that random selection for pruning ResNet-32 on CIFAR-10/CIFAR-100 shows good results, even better than magnitude-based methods in some cases. However, the performance of random pruning is not robust. As shown in Fig.~\ref{Fig.5}(b) and Fig.~\ref{Fig.5}(d), randomly selecting channels can lead to bad results, especially when we prune more channels and the percentage of the remained FLOPs is lower than 40\%. 

\subsection{Ablation Study}
\label{ablationstudy}
\subsubsection{Normalization Strategies} 
We conduct experiments on pruning ResNet-32/ResNet-56/VGG-16 on CIFAR-10 to compare two normalization strategies utilized for pruning less important channels with high memory costs and non-normalization. As shown in Table~\ref{ablation-ResNet32-on-CIFAR10-Norm}-\ref{ablation-ResNet-56-on-CIFAR10-Norm}, the two normalization strategies are denoted by ``IFSO-'' and ``Norm-'' prefixes, where the raw scores computed by Eq.\ref{eq:sensitivity_score} are normalized by $\sqrt{\Delta Mem}$ and by $\Delta Mem$, respectively. 
The experimental results show that ``IFSO-'' performs best under the same number of FLOPs except in rare cases. Furthermore, compared with the “Norm-” strategy, ``IFSO-'' consistently achieves higher Top-1 accuracy under the same FLOPs but with a higher prune percentage of weights. 

Besides, we observe an interesting phenomenon: non-normalization (i.e., pruning models with raw scores) generally removes more than twice as many weights as that by the ``Norm-'' strategy under the same number of FLOPs. 
We investigate the reason for the significant difference in weight percentage. For ResNet and VGG, the reduction of memory caused by pruning one channel is higher in shallow layers due to the larger input feature maps. However, the reduction of weights is more considerable in deeper layers. We notice that the models pruned with non-normalization have the typical behavior (\cite{su2020sanity,wang2020picking}) of keeping less capacity in deeper layers. In contrast, the model pruned by ``Norm-'' prefers to keep more capacity in later stages by suppressing the channel scores in shallow layers via their bigger $\Delta Mem$.

Although non-normalization keeps only half as many weights as ``Norm-'' or even less, the models pruned with non-normalization achieve better Top-1 accuracy on ResNet-56 and VGG-16. 
However, for ResNet-32, non-normalization does not compete with ``Norm-''. It is possible because ResNet-32 is a relatively small and compact model with fewer redundant weights than ResNet-56/VGG-16. Therefore, a rapid decline in the number of weights may result in a relatively faster drop in accuracy. However, the ``IFSO-'' strategy can alleviate the sharp drop in the number of weights and achieve competitive accuracy.  



\begin{table}[t]
  \caption{Performance of pruning ResNet-32 on CIFAR-10 with different normalization policies.}
  \label{ablation-ResNet32-on-CIFAR10-Norm}
  \centering
  \begin{tabular}{l|cc|cccc}
    \Xhline{0.3ex}
    \multirow{2}{*}{Method} & FLOPs $\downarrow$ & $W$ & Original & Pruned& T1 $\downarrow$ \\
    & (\%) & (\%) & T1(\%) & T1(\%) & (\%) \\
    \hline 
    Norm-40 & 40.0 & 86.43 & 93.51 & \underline{93.11} & \underline{0.40} \\
    Raw-40 & 40.0 & 43.07 & 93.51 & 92.83 & 0.68 \\
    IFSO-40 (\textbf{ours}) & 40.0 & 67.51 & 93.51 & \textbf{93.23} & \textbf{0.28} \\
    \hline
    Norm-50 & 50.0 & 79.90 & 93.51 & \underline{92.57} & \underline{0.94} \\
    Raw-50 & 50.0 & 31.91 & 93.51 & 92.42 & 1.09 \\
    IFSO-50 (\textbf{ours}) & 50.0 & 57.31 & 93.51 & \textbf{93.00} & \textbf{0.51} \\
    \Xhline{0.3ex}
  \end{tabular}
  \flushleft{``$W$'' stands for the \textbf{remaining} weights' percentage.
  }
\end{table}

\begin{table}[t]
  \caption{Performance of pruning VGG-16 on CIFAR-10 with different normalization policies.}
  \label{ablation-VGG-16-on-CIFAR10-Norm}
  \centering
  \begin{tabular}{l|cc|ccc}
    \Xhline{0.3ex}
    \multirow{2}{*}{Method} & FLOPs $\downarrow$ & $W$ & Original & Pruned& T1 $\downarrow$ \\
    & (\%) & (\%) & T1(\%) & T1(\%) & (\%) \\
    \hline 
    Norm-40 & 40.0 & 46.85 & 93.34 & 93.56 & -0.22 \\
    Raw-40 & 40.0 & 20.63 & 93.34 & \underline{93.64} & \underline{-0.30} \\
    IFSO-40 (\textbf{ours}) & 40.0 & 27.16 & 93.34 & \textbf{94.04} & \textbf{-0.70} \\
    \hline
    Norm-50 & 50.0 & 35.01 & 93.34 & 93.61 & -0.27 \\
    Raw-50 & 50.0 & 12.07 & 93.34 & \underline{93.78} & \underline{-0.44} \\
    IFSO-50 (\textbf{ours}) & 50.0 & 17.03 & 93.34 & \textbf{93.94} & \textbf{-0.60} \\
    \Xhline{0.3ex}
  \end{tabular}
  \flushleft{``$W$'' have the same meaning as Table~\ref{ablation-ResNet32-on-CIFAR10-Norm}.} 
\end{table}

\begin{table}[t]
  \caption{Performance of pruning ResNet-56 on CIFAR-10 with different normalization policies.}
  \label{ablation-ResNet-56-on-CIFAR10-Norm}
  \centering
  \begin{tabular}{l|cc|ccc}
    \Xhline{0.3ex}
    \multirow{2}{*}{Method} & FLOPs $\downarrow$ & $W$ & Original & Pruned& T1 $\downarrow$ \\
    & (\%) & (\%) & T1(\%) & T1(\%) & (\%) \\
    \hline 
    Norm-40 & 40.0 & 81.54 & 93.68 & 93.79 & -0.11 \\
    Raw-40 & 40.0 & 45.61 & 93.68 & \underline{93.94} & \underline{-0.26} \\
    IFSO-40 (\textbf{ours}) & 40.0 & 65.97 & 93.68 & \textbf{93.95} & \textbf{-0.27} \\
    \hline
    Norm-50 & 50.0 & 74.88 & 93.68 & 93.28 & 0.40 \\
    Raw-50 & 50.0 & 34.62 & 93.68 & \textbf{93.72} & \textbf{-0.04} \\
    IFSO-50 (\textbf{ours}) & 50.0 & 56.23 & 93.68 & \underline{93.65} & \underline{0.03} \\
    \Xhline{0.3ex}
  \end{tabular}
  \flushleft{``$W$'' has the same meaning as Table~\ref{ablation-ResNet32-on-CIFAR10-Norm}.}
\end{table}

\subsubsection{Approximation of $H^{-1}$}
\label{approximation_of_inverse_H}
Apart from using identity to approximate $H^{-1}$ in Eq.\eqref{eq:sensitivity_score}, we also explore two commonly used methods of computing the Inverse-Hessian Vector Products ($H^{-1}v$): Neumann series \cite{lorraine2020optimizing} and Sherman-Morison approximation \cite{singh2020woodfisher}, and we empirically found Neumann series achieves better results than Sherman-Morison approximation. In contrast, identity approximation obtains the best results among the three methods. Specially, we compare identity approximation (IFSO) with Neumann approximations under different approximation terms when pruning ResNet-32 on CIFAR-10. The results are shown in Table~\ref{ablation-ResNet32-on-CIFAR10-neumann}, where Neumann-1 and Neumann-2 have different expansion terms, as described in the notes under this table. We can see that IFSO can achieve higher Top-1 accuracy under the same number of FLOPs, while Neumann-2 is the worst. The results suggest that the identity matrix is more suitable for computing our proposed channel importance score.



\begin{table}[t]
  \caption{Performance of pruning ResNet-32 on CIFAR-10 with different Neumann approximations.}
  \label{ablation-ResNet32-on-CIFAR10-neumann}
  {\centering
    \begin{tabular}{l|c|ccc}
    \Xhline{0.3ex}
    \multirow{2}{*}{Method} & FLOPs $\downarrow$ & Original & Pruned& T1 $\downarrow$ \\
    & (\%) & T1(\%) & T1(\%) & (\%) \\
    \hline 
    Neumann-1-40 & 40.0 & 93.51 & \underline{93.14} & \underline{0.37} \\
    Neumann-2-40 & 40.0 & 93.51 & 93.05 & 0.46 \\
    IFSO-40 (\textbf{ours}) & 40.0 & 93.51 & \textbf{93.23} & \textbf{0.28} \\
    \hline
    Neumann-1-50 & 50.0 & 93.51 & \underline{92.73} & \underline{0.78} \\
    Neumann-2-50 & 50.0 & 93.51 & 92.67 & 0.84 \\
    IFSO-50 (\textbf{ours}) & 50.0 & 93.51 & \textbf{93.00} & \textbf{0.51} \\
    \Xhline{0.3ex}
  \end{tabular}
  \flushleft{The row with ``IFSO-'' prefix means $H^{-1}$ is replaced with identity, ``Neumann-1-'' applies $H^{-1}v=V_0+V_1$, ``Neumann-2-'' uses $H^{-1}v=V_0+V_1+V_2$, where $V_0=v$, $V_k=(I-\gamma H)V_{k-1}$, $\gamma$ is a small enough scalar.}}
\end{table}

\begin{figure*}[t]
\centering
  \subfloat[Prune ResNet-32]{
  \begin{minipage}{4.2cm}
      \includegraphics[width=4.2cm,height=3.5cm]{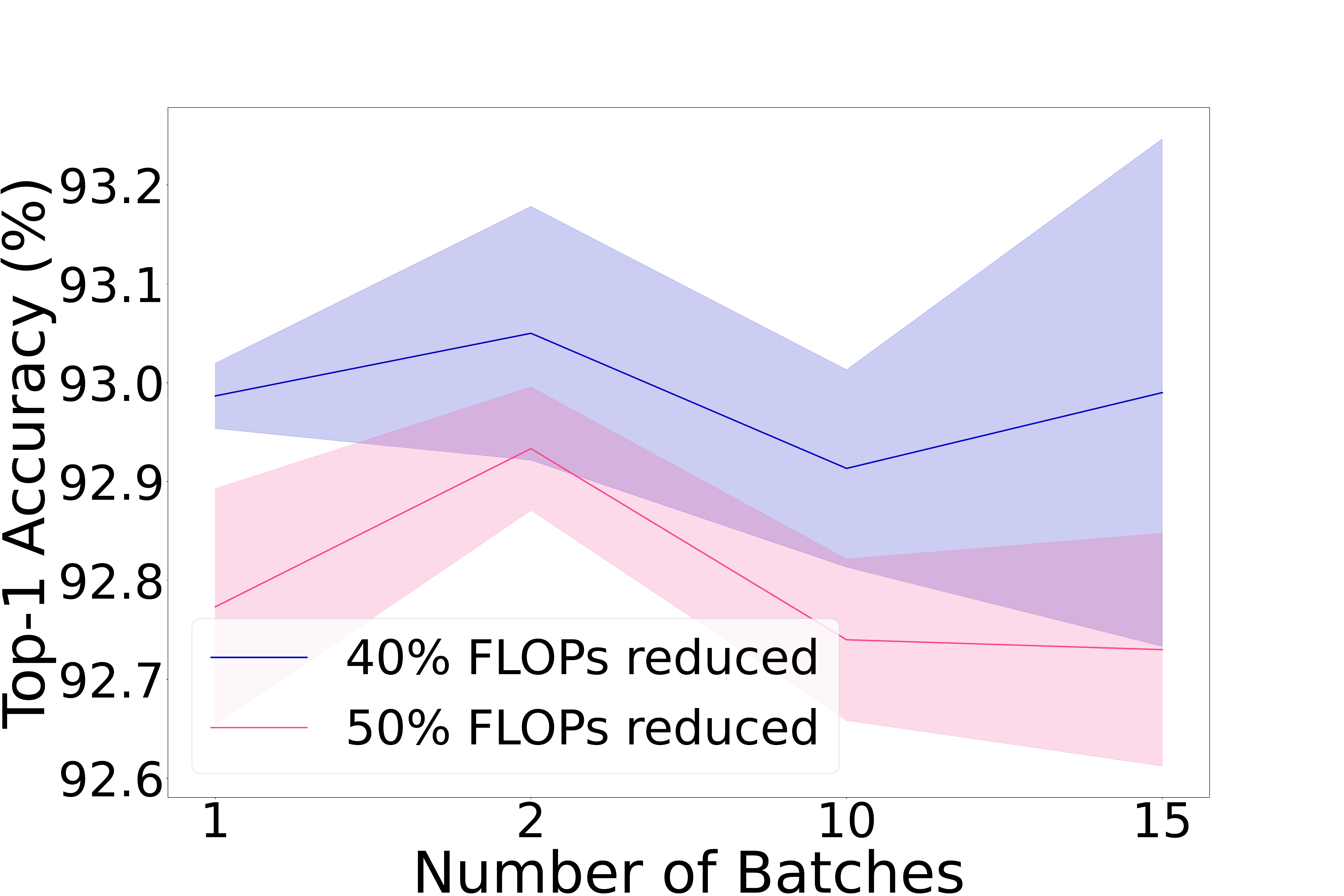}
        \label{fig2a:ResNet-32}
  \end{minipage}}
  \subfloat[Prune ResNet-56]{
  \begin{minipage}{4.2cm}
       \includegraphics[width=4.2cm,height=3.5cm]{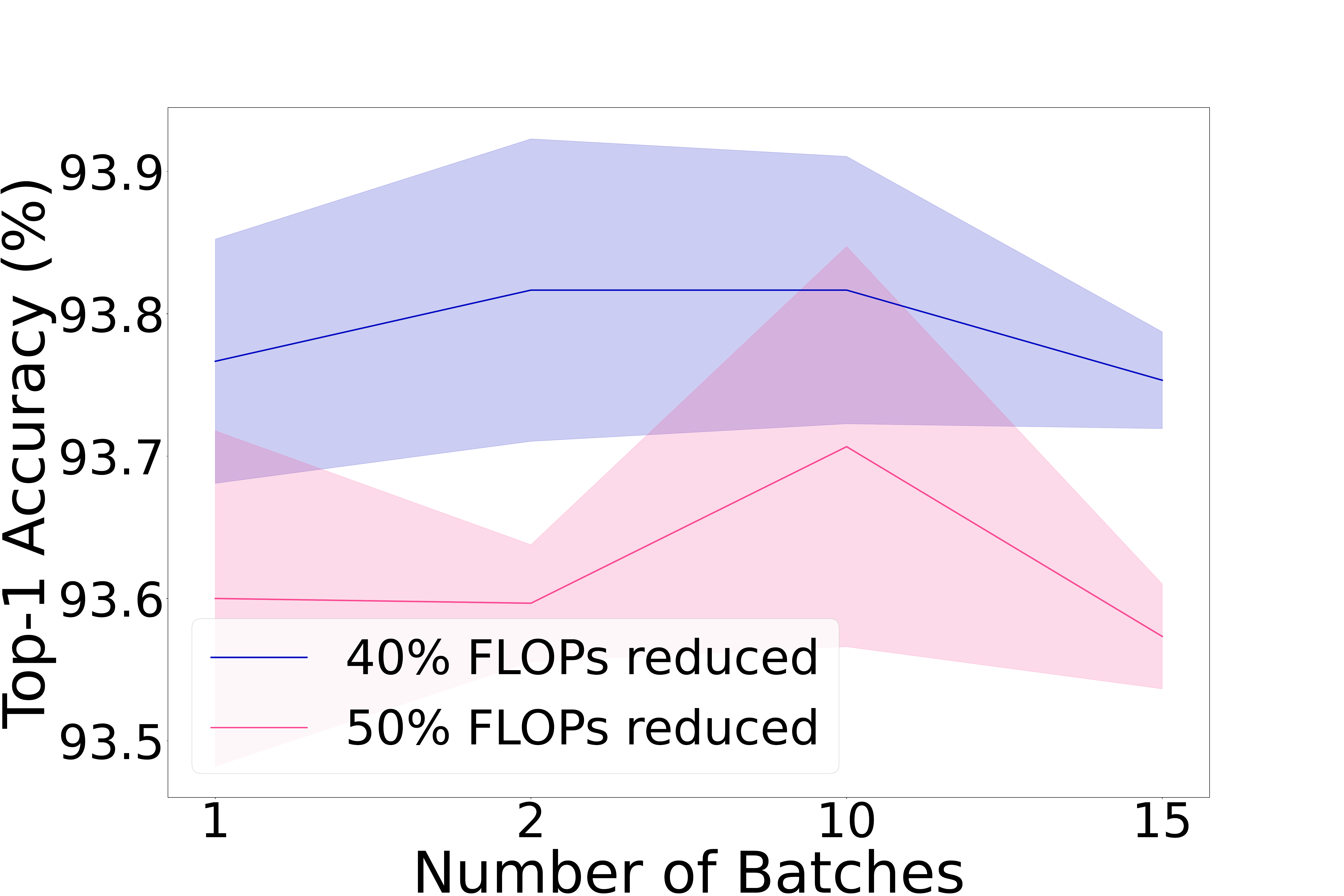}
         \label{fig2b:ResNet-56}
  \end{minipage} 
  }
  \subfloat[Prune VGG-16]{
  \begin{minipage}{4.2cm}
       \includegraphics[width=4.2cm,height=3.5cm]{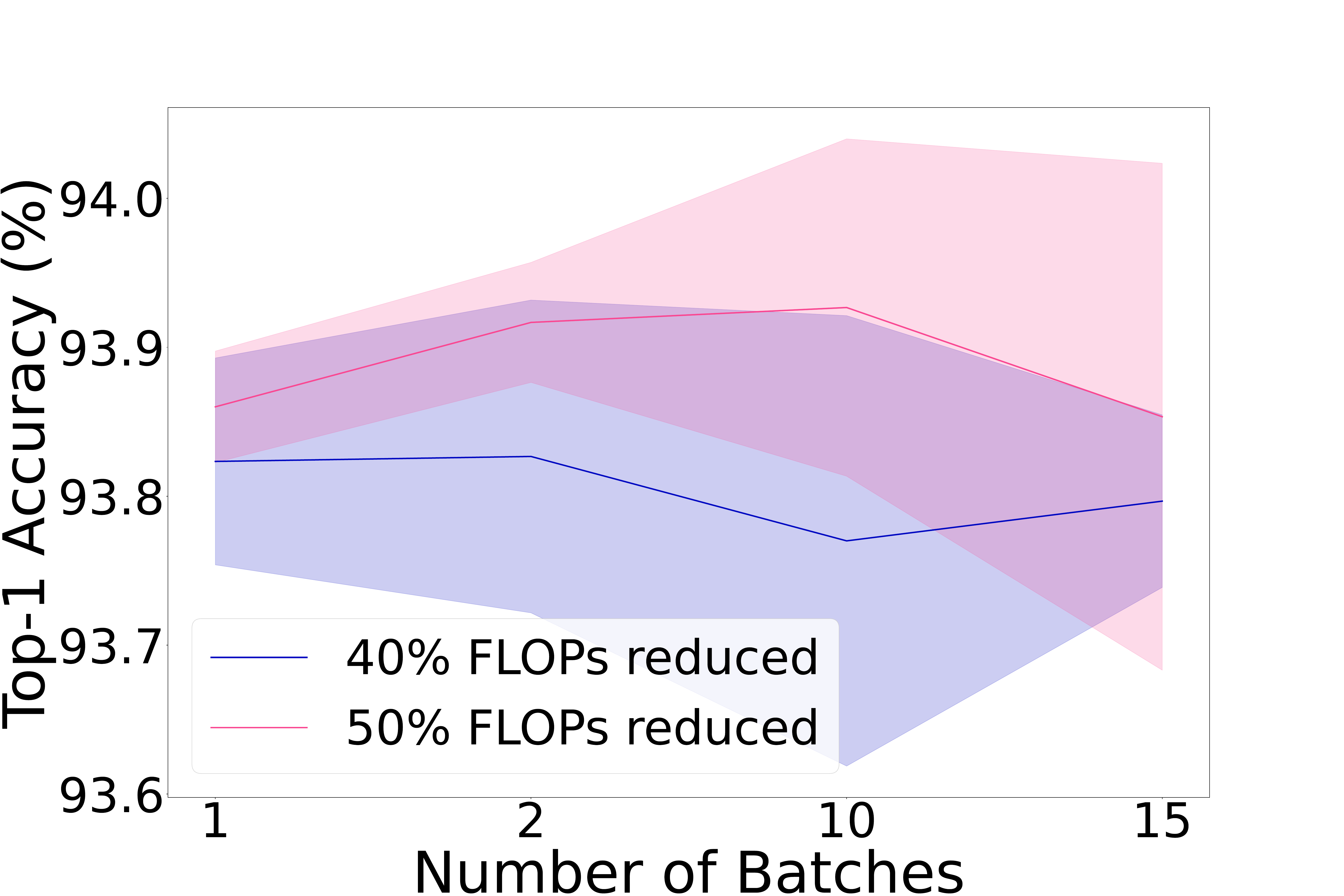}
         \label{fig2c:VGG-16}
  \end{minipage} 
  }  
 \caption{The influence of the number of batches of the proxy dataset on performance on CIFAR-10. Shaded regions indicate standard deviation.}
 \label{Fig.6}
\vspace{-0.3cm}
\end{figure*}

\subsubsection{Number of Batches in Proxy Dataset} 
In our main experiments, we average channel scores for efficiency under two batches of training datasets (used as the proxy dataset). To study the effect of using different numbers of batches on performance, we use 1, 2, 10, and 15 batches of a proxy dataset to calculate channel sensitivity scores by Eq.\eqref{eq:sensitivity_score}, respectively. We conduct this experiment under different prune percentages (40\% and 50\% FLOPs reduction) for ResNet and VGGNet on CIFAR-10. 
Experimental results are the mean and standard deviation of Top-1 accuracy over three independent runs, as shown in Figure~\ref{Fig.6}. 
We observe that more batches of the proxy dataset are not required for better performance and likely occur larger accuracy oscillation. 

In addition, fewer batches are used, and less time is consumed.
By using two batches of training datasets, an excellent balance between performance and time consumption can be achieved, and we thus follow this setting to evaluate channel scores in our main experiments.

\subsubsection{Different Accumulated Times for Channel Scores}
\label{pruneinterval}
\begin{figure}[H]
\centering
\begin{minipage}{7.5cm}
\includegraphics[width=7.5cm,height=5cm]{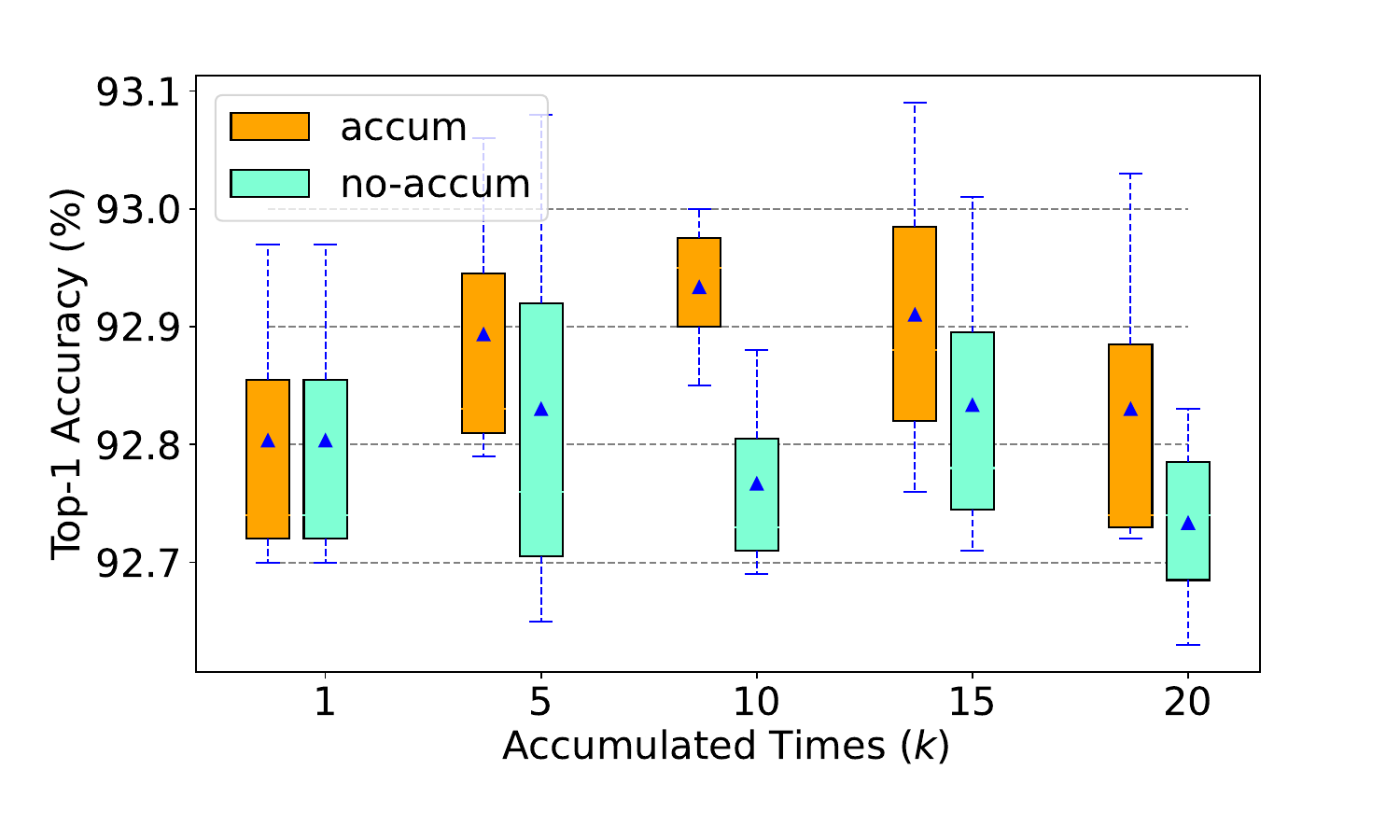}
\end{minipage}
\caption{Different accumulated times for channel scores on ResNet-32 on CIFAR-10 with 50\% FLOPs reduced. The little triangle in each box indicates the mean value of Top-1 accuracy. We choose $k$ to be 10, where the highest mean value is obtained.}
\label{Fig.8}
\end{figure}

As mentioned in Algorithm~\ref{alg2}, we perform one pruning action when channel scores are accumulated $k$ times. To study the effect of using different accumulated times $k$ on channel scores, we set $k$ to 1, 5, 10, 15, or 20 when we prune 50\% FLOPs from ResNet-32 on CIFAR-10. We consider two cases for a specific $k$ to get channel scores. One is to accumulate channel scores $k$ times before each pruning action, and another is to use the $k$-th values as the scores. The experimental results are illustrated in Fig.\ref{Fig.8}, where the values of each box are based on three independent runs. The Top-1 accuracy of each box (denoted as ``accum'') and box (denoted as ``no-accum'') is based on the pruned model using the accumulated scores or the $k$-th scores to prune, respectively. 

We can see from Fig.\ref{Fig.8} that, for each $k$, the accumulated scores always yield higher average results of Top-1 accuracy than using the $k$-th scores, demonstrating the superiority of the accumulation scheme. These results imply that multiple accumulations of scores may be similar to multiple sampling, which helps the scores more robust. Based on our ablation study on $k$, our IFSO achieves the highest Top-1 accuracy when $k$ is 10, and we thus set $k$ to 10 in our main experiments unless otherwise specified. 

\subsubsection{Combination of One-shot and Incremental Pruning}
\label{combinationofprunings}
As mentioned in Section \ref{channelsensitivity}, our score $S$ in Eq.\eqref{eq:sensitivity_score} can be used for both one-shot and incremental pruning. Generally, the former is more efficient than the latter, but with the cost of performance degeneration. One way to balance performance and efficiency is to combine one-shot and incremental pruning. Given a FLOPs reduction percentage $p$, we study Top-1 accuracy and pruning time-consuming across different combination percentages of single-pass (i.e., IFSO-one-shot) and incremental-way (i.e., IFSO). We define the percentage of FLOPs pruned by IFSO: $\frac{t}{p}\times 100\%$, where $t\in[0,p]$, $t$ stands for $t$ FLOPs are reduced by IFSO, and then $(p-t)$ FLOPs are reduced by IFSO-one-shot. 

\begin{figure}[t]
\centering
\begin{minipage}{7.5cm}
\includegraphics[width=7.5cm,height=5cm]{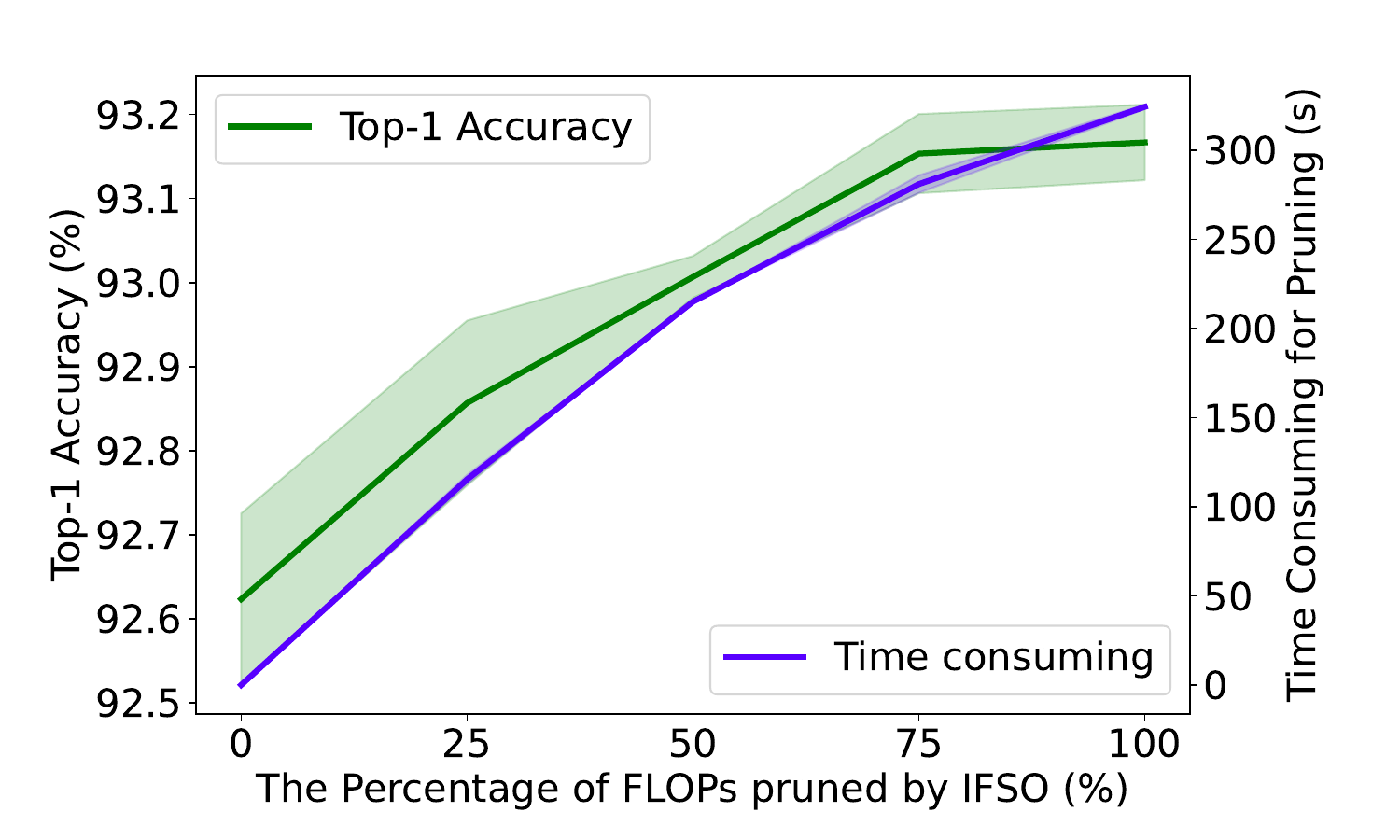}
\end{minipage}
\caption{Combination of IFSO-one-shot and IFSO (incremental-way) for ResNet-32 on CIFAR-10 with a total 40\% FLOPs reduction. A suitable combination of IFSO-one-shot and IFSO achieves a better balance between accuracy and pruning efficiency. Shaded regions indicate standard deviation.}
\label{Fig.7}
\end{figure}

We take $p=40\%$ as an example for pruning ResNet-32 on CIFAR-10 over three independent runs. As illustrated in Fig.~\ref{Fig.7}, as more FLOPs are pruned through IFSO, Top-1 accuracy and time consumption increase. For example, under the equivalent sparsity constraint, we improve Top-1 accuracy of the pruned models from 92.48\% (The target FLOPs constraint is achieved by one-shot.) to 93.23\% (The target FLOPs budget is approached gradually.) at the cost of about 320 seconds. Therefore, when the pruning time cost is affordable, pruning models by IFSO can lead to a better performance than IFSO-one-shot. However, we also observe no significant improvement in Top-1 accuracy when the percentage of FLOPs pruned by IFSO is from 75\% to 100\%. On the whole, the phenomenon implies that a suitable mixture of one-shot and incremental pruning schemes can maintain performance while improving pruning efficiency.

\section{Conclusion}
Minimal performance degeneration, e.g., loss changes before and after pruning being smallest is widely used in pruning. To accurately evaluate the truth performance degeneration requires retraining the survived weights, which is prohibitively slow. Hence existing pruning methods use previous weights to evaluate the performance degeneration. We observe that with and without retraining, the loss changes differ significantly. This makes us wonder if we can accurately estimate the true loss change without retraining. Inspired by influence functions, we derive a closed-form estimator for the true loss change per mask change. We then show how to assess the importance of all channels simultaneously and develop a novel global channel pruning algorithm accordingly. Extensive experiments on classic CNNs for both image classification and object detection tasks have demonstrated the effectiveness of the proposed method over the state-of-the-art channel pruning methods. To the best of our knowledge, we are the first that shows evaluating true loss changes for pruning without retraining is possible. This work will open up new opportunities for a series of future works to emerge.

{\small
\bibliographystyle{IEEEtranN}
\bibliography{ref}
}
\ifCLASSOPTIONcaptionsoff
  \newpage
\fi

\end{document}


%
\title{Influence Function Based Second-Order Channel Pruning: Evaluating True Loss Changes For Pruning Is Possible Without Retraining}
%
%
%
%

\author{Hongrong~Cheng*,~Miao Zhang*$\dagger$,~\IEEEmembership{Member,~IEEE},
Javen~Qinfeng~Shi,~\IEEEmembership{Member,~IEEE}
\IEEEcompsocitemizethanks{
\IEEEcompsocthanksitem *Equal contribution. \\
H. Cheng and J. Q. Shi are with the University of Adelaide. M. Zhang is with Harbin Institute of Technology (Shenzhen). \\
E-mail:\{hongrong.cheng, javen.shi\}@adelaide.edu.au, 
\{zhangmiao\}@hit.edu.cn.\\
$\dagger$ Corresponding author.
}
\thanks{Preprint. Under review}}

\maketitle

\IEEEdisplaynontitleabstractindextext

%
\IEEEpeerreviewmaketitle

\ifCLASSOPTIONcompsoc
\IEEEraisesectionheading{\section*{Proof}\label{sec:proof}}
\else
\section*{Proof}
\label{sec:proof}
\fi

\begin{prop}
\label{prop1}
Suppose that the deep neural network model obtains the optimized weights $W^*$ with mask $M$ after training it to converge, mask $M$ changes to $\hat{M}$ after removing some specific channels, and the validation loss for $\hat{M}$ is $L(\hat{W}^*,\hat{M})$. If the third and higher derivatives of the loss function \textit{w.r.t.} weights at optimum are zero or sufficiently small, and with $\frac{\partial L(\hat{W^*},\hat{M})}{\partial W} = 0$, we have
\begin{equation}
\begin{aligned}
    \Delta L_{gt} &= L(\hat{W}^*,\hat{M}) - L(W^*,M) \\
    &= \Delta L_{ex}-\frac{1}{2} \cdot \frac{\partial L(W^*,\hat{M})}{\partial W}^{T} \cdot \frac{\partial^2L(W^*,\hat{M})^{-1}}{\partial W \partial W} \cdot  \frac{\partial L(W^*,\hat{M})}{\partial W},
\label{eq:prop_1}
\end{aligned}
\end{equation}
where $\Delta L_{ex} = L(W^*, \hat{M})-L(W^*,M)$.
\end{prop}

\newenvironment{proof_p1}{\noindent\textbf{\textit{Proof of Proposition \ref{prop1}:}}}{  $\square$}
\begin{proof_p1}
Following influence functions \cite{basu2020on,koh2017understanding}, we conduct a second-order Taylor expansion on $\hat{W}^*$ under the assumption that the third and higher derivatives of the loss function at optimum is sufficiently small \cite{basu2020on}, the ground truth loss change $\Delta L_{gt}$ can be estimated by:
\begin{equation}
\begin{aligned}
\Delta L_{gt} &= L(\hat{W}^*,\hat{M}) - L(W^*,M) \\
& = L(W^*,\hat{M}) + \Delta W^{T} \cdot \frac{\partial L(W^*,\hat{M})}{\partial W} \\
& + \frac{1}{2} \cdot \Delta W^{T} \cdot \frac{\partial^2L(W^*,\hat{M})}{\partial W \partial W} \cdot \Delta W - L(W^*,M),
\end{aligned}
\label{eq.prop_1_step2}
\end{equation}
where $\Delta W = \hat{W}^* - W^*$.

As the model is supposed to be converged after retraining, we have $\frac{\partial L(\hat{W^*},\hat{M})}{\partial W} = 0$. In this part, we consider the first-order Taylor expansion on $\hat{W}^*$ and have
\begin{equation}
\begin{aligned}
\frac{\partial L(\hat{W^*},\hat{M})}{\partial W} &= \frac{\partial L(W^*,\hat{M}) + \Delta W^{T} \cdot \frac{\partial L(W^*,\hat{M})}{\partial W}}{\partial W} \approx 0.
\label{eq.prop_1_step3}
\end{aligned}
\end{equation}

\begin{equation}
\frac{\partial L(W^*,\hat{M})}{\partial W}+ \frac{\partial^{2}L(W^*,\hat{M})}{\partial W \partial W} \cdot \Delta W \approx 0
\end{equation}

Then we have
\begin{equation}
    \Delta W =- \frac{\partial^2L(W^*,\hat{M})^{-1}}{\partial W \partial W} \cdot \frac{\partial L(W^*,\hat{M})}{\partial W}.
\label{eq.prop_1_step4}    
\end{equation}
Substituting Eq.\eqref{eq.prop_1_step4} into Eq.\eqref{eq.prop_1_step2}, we get
\begin{equation}
\begin{aligned}
    \Delta L_{gt} &= \Delta L_{ex} \\
    & - \frac{1}{2} \cdot \frac{\partial L(W^*,\hat{M})}{\partial W}^{T} \cdot \frac{\partial^2L(W^*,\hat{M})^{-1}}{\partial W \partial W} \cdot \frac{\partial L(W^*,\hat{M})}{\partial W}.
\end{aligned}
\label{eq:prop_1_step5}
\end{equation}
\end{proof_p1}

\begin{prop}
\label{prop2}
Suppose the channel mask $M$ is continuous, $\Delta M=\hat{M} - M$ is infinitesimally small. Assume that the third and higher order derivatives of the loss function \textit{w.r.t.} $W$ at optimum are sufficiently small. Given a trained network model with the optimized weighs $W^*$ on the mask $M$, the sensitivity of $\Delta L_{gt}$ \textit{w.r.t.} $\Delta M$ can be estimated as:
\begin{equation}
    \left |\frac{\Delta L_{gt}}{\Delta M}\right |\approx\left |\frac{1}{2} \cdot \Delta M^{T} \cdot \frac{\partial^2 L(W^*,M)}{\partial W \partial M} \cdot H^{-1} \cdot \frac{\partial^2 L(W^*,M)}{\partial M \partial W} \right |,
\label{eq:prop_2}
\end{equation}
where $H=\frac{\partial^2 L(W^*,M)}{\partial W \partial W}$ is the Hessian matrix. 
\end{prop}
\newenvironment{proof_p2}{\noindent\textbf{\textit{Proof of Proposition \ref{prop2}:}}}{$\square$}
\begin{proof_p2}
Since the change on the $M$ is infinitesimal, we can now conduct first-order Taylor expansion on $M$ for $\Delta L_{gt}$ as: 
\begin{equation}
\begin{aligned}
\Delta L_{gt} & = L(\hat{W}^*,\hat{M}) - L(W^*,M) = L(\hat{W}^*,M) \\
& + \Delta M^{T} \cdot \frac{\partial L(\hat{W}^*,M)}{\partial M} - L(W^*,M).
\end{aligned}
\label{eq:prop_2_step1}
\end{equation}

Then, we conduct a second-order Taylor expansion on $\hat{W}^*$ of $L(\hat{W}^*,M)$ in Eq.\eqref{eq:prop_2_step1}, and obtain
\begin{equation}
\begin{aligned}
L(\hat{W}^*,M) & = L(W^*,M) + \Delta W^{T} \cdot \frac{\partial L(W^*,M)}{\partial W} \\
& + \frac{1}{2} \cdot \Delta W^{T} \cdot \frac{\partial^2L(W^*,M)}{\partial W \partial W} \cdot \Delta W.
\end{aligned}
\label{eq:prop_2_step2}    
\end{equation}

As the model is optimized based on channel mask $M$, we have $\frac{\partial L(W^*,M)}{\partial W} = 0$. Then Eq.\eqref{eq:prop_2_step2} is simplified as
\begin{equation}
L(\hat{W}^*,M) = L(W^*,M) + \frac{1}{2} \cdot \Delta W^{T} \cdot \frac{\partial^2L(W^*,M)}{\partial W \partial W} \cdot \Delta W.
\label{eq:prop_2_step3}
\end{equation}

Substituting Eq.\eqref{eq:prop_2_step3} into Eq.\eqref{eq:prop_2_step1}, we have
\begin{equation}
\Delta L_{gt} =  \frac{1}{2} \cdot \Delta W^{T} \cdot \frac{\partial^2L(W^*,M)}{\partial W \partial W} \cdot \Delta W+\Delta M^{T} \cdot \frac{\partial L(\hat{W}^*,M)}{\partial M}. 
\label{eq:prop_2_step4}
\end{equation}

When assuming the third and higher derivatives of the loss function $L$ \textit{w.r.t.} model weights $W$ at optimum are sufficiently small, we can further conduct Taylor expansion on $\hat{W}^*$ in Eq.\eqref{eq:prop_2_step4} and have 
\begin{equation}
\begin{aligned}
\Delta L_{gt} &=  \frac{1}{2} \cdot \Delta W^{T} \cdot \frac{\partial^2L(W^*,M)}{\partial W \partial W} \cdot \Delta W \\
& + \Delta M^{T} \cdot \frac{L(W^*,M) + \Delta W^{T} \cdot \frac{\partial L(W^*,M)}{\partial W}}{\partial M}. 
\end{aligned}
\label{eq:prop_2_step5}      
\end{equation}
As the weights $W^*$ with the mask $M$ are optimized, we have $\frac{\partial L(W^*,M)}{\partial M}\approx0$. Then Eq.\eqref{eq:prop_2_step5} is simplified as
\begin{equation}
\begin{aligned}
\Delta L_{gt} & \approx \frac{1}{2} \cdot \Delta W^{T} \cdot \frac{\partial^2L(W^*,M)}{\partial W \partial W} \cdot \Delta W \\
& + \Delta M^{T} \cdot \frac{\partial^2 L(W^*,M)}{\partial M {\partial W}} \cdot \Delta W. 
\end{aligned}
\label{eq:prop_2_step6}      
\end{equation}

To reveal the relationship between $\Delta L_{gt}$ and $\Delta M$, we need to further approximate $\Delta W$. Similar as before, we utilize $\frac{\partial L(\hat{W}^*,\hat{M})}{\partial W}=0$ and Taylor expansion on $\hat{W}^*$ (omitting third and higher derivatives), and we have
\begin{equation}
\frac{\partial L(\hat{W}^*,\hat{M})}{\partial W} = \frac{\partial{L(W^*,\hat{M})} + \Delta W^{T} \cdot \frac{L(W^*, \hat{M})}{\partial W}}{\partial W}=0.  
\label{eq:prop_2_step7}
\end{equation}
For $L(W^*,\hat{M})$ in Eq.\eqref{eq:prop_2_step7}, we consider first-order Taylor expansion on $\hat{M}$ since the change on the $M$ is infinitesimal, and have 
\begin{equation}
L(W^*,\hat{M}) = L(W^*, M) + \Delta M^{T} \cdot \frac{\partial{L(W^*, M)}}{\partial M}. 
\label{eq:prop_2_step8}
\end{equation}
Substituting Eq.\eqref{eq:prop_2_step8} into Eq.\eqref{eq:prop_2_step7}, we get
\begin{equation}
\frac{L(W^*, M) + \Delta M^{T} \cdot \frac{\partial{L(W^*, M)}}{\partial M} + \Delta W^{T} \cdot \frac{L(W^*,\hat{M})}{\partial W}}{\partial W} = 0.
\label{eq:prop_2_step9}
\end{equation}

Similar as before, we have $\frac{\partial L(W^*,M)}{\partial W}=0$ and $\frac{\partial L(W^*,M)}{\partial M}\approx 0$ based on the local optimum. Based on the first-order Taylor expansion on $\hat{M}$ and neglecting third and higher derivatives, we have
\begin{equation}
\Delta M^{T} \cdot \frac{\partial^2{L(W^*, M)}}{\partial M \partial W}+\Delta W^{T} \cdot \frac{\partial^2 L(W^*,M)}{\partial W \partial W}\approx 0.
\label{eq:prop_2_step10}
\end{equation}
So,
\begin{equation}
\Delta W^{T} \approx -\Delta M^{T} \cdot \frac{\partial^2 L(W^*,M)}{\partial M \partial W} \cdot \frac{\partial^2 L(W^*,M)^{-1}}{\partial W \partial W}.    
\label{eq:prop_2_step11}
\end{equation}

Substituting Eq.\eqref{eq:prop_2_step11} into Eq.\eqref{eq:prop_2_step4}, we can get 
\begin{equation}
\Delta L_{gt}\approx-\frac{1}{2} \cdot \Delta M^{T} \cdot \frac{\partial^2 L(W^*,M)}{\partial W \partial M} \cdot H^{-1} \cdot \frac{\partial^2 L(W^*,M)}{\partial M \partial W} \cdot \Delta M,
\label{eq:prop_2_step12}
\end{equation}
where $H=\frac{\partial^2 L(W^*,M)}{\partial W \partial W}$ is the Hessian matrix. 
From Eq.\eqref{eq:prop_2_step12}, we can observe the relationship between $\Delta L_{gt}$ and $\Delta M$. The the sensitivity of $\Delta L_{gt}$ \textit{w.r.t.} $\Delta M$ can be estimated as:
\begin{equation}
    \left |\frac{\Delta L_{gt}}{\Delta M}\right |\approx\left |\frac{1}{2} \cdot \Delta M^{T} \cdot \frac{\partial^2 L(W^*,M)}{\partial W \partial M} \cdot H^{-1} \cdot \frac{\partial^2 L(W^*,M)}{\partial M \partial W} \right |.
\end{equation}

Therefore, Proposition \ref{prop2} is proved. 

\end{proof_p2}

\begin{corollary}
\label{corollary1}
Based on the Assumption \ref{assumption1}-\ref{assumption2}, we could bound the error between the approximated validation loss $\mathcal{L}(\hat{W}^*, \hat{M})$ and the ground-truth $\tilde{\mathcal{L}}(\hat{W}^*, \hat{M})$ with $E =\left \|\mathcal{L}(\hat{W}^*, \hat{M}) -\tilde{\mathcal{L}}(\hat{W}^*, \hat{M})  \right \| \leq \frac{\left | \Delta W \right |^3}{6} \textup{max}\left | \frac{\partial \mathcal{L}^3}{\partial W^3} \right |$,
where $\left \| \Delta W \right \| \leq \frac{C_{l}}{\lambda} + \frac{C_H \cdot C_l^2}{2\sigma_{min}^2\lambda}$, $\sigma_{min}$ is the smallest eigenvalue of Hessian matrix $\frac{\partial^2 \mathcal{L}({W}^*, \hat{M})}{\partial W \partial W}$.
\end{corollary}

\newenvironment{proof_c1}{\noindent\textbf{\textit{Proof of Corollary \ref{corollary1}:}}}{  $\square$}
\begin{proof_c1}
To analyze the error bound of approximation in Proposition \ref{prop1}, we need the following assumptions which are commonly used in the bi-level optimization (\cite{couellan2016convergence,zhang2021idarts}). 
\begin{assumption}
\label{assumption1}
$\mathcal{L}(W, M)$ is twice differentiable with constant $C_H$ and is $\lambda$-strongly convex with $W$ around $W^*(M)$.  
\end{assumption}
\begin{assumption}
\label{assumption2}
The $\left\|\frac{\partial L(W^*, \hat{M})}{\partial W}\right\|$ is bounded with constant $C_{l}>0$.
\end{assumption}

Now we can estimate the error bound in Proposition \ref{prop1}. First, due to the second-order Taylor expansion on $W$ in Eq.\eqref{eq.prop_1_step2}, we have the error
\begin{equation}
\label{eq:error_bound}
E=\left \| \mathcal{L}(\hat{W}^*, \hat{M}) - \tilde{\mathcal{L}}(\hat{W}^*, \hat{M})  \right \| \leq \frac{\left | \Delta W \right |^3}{6} \textup{max}\left | \frac{\partial \mathcal{L}^3}{\partial W^3} \right |,
\end{equation} 
where the approximated validation loss $\mathcal{L}(\hat{W}^*, \hat{M})$ is Taylor expansion of $\hat{\mathcal{L}}$ in the point $\hat{W}^*-\Delta W$. Since $\Delta W$ in Eq.\eqref{eq.prop_1_step4} is an approximation, $W^*+\Delta W\neq\hat{W}^*$, where $\hat{W}^*$ is the true optimal with the mask $\hat{M}$. We define the approximate local optimal with the mask $\hat{W}^*$ as $\hat{W}^*_e:=W^*+\Delta W$ and $\Delta W_e = \hat{W}^*-\hat{W}_e^*$.

First, we have
\begin{equation} 
\begin{aligned}
\left \| \Delta W \right \|  =\left \| W^*-\hat{W}^*+\hat{W}^*-\hat{W}_e^* \right \| \leq \left \| W^*-\hat{W}^*\right \|+\left \|\hat{W}^*-\hat{W}_e^* \right \|.
\end{aligned}
\label{eq:theta_two_part}
\end{equation}

For the first term in Eq.\eqref{eq:theta_two_part}, based on Assumption \ref{assumption1} that $\mathcal{L}(W, M)$ is $\lambda$-strongly convex with $W$ around $W^*$, we have:
\begin{equation} 
\label{eq.first_term}
\left \| \nabla \mathcal{L}(\hat{W}^*, \hat{M})- \nabla \mathcal{L}(W^*, \hat{M}) \right \| \geq \lambda \cdot \left \| \hat{W}^*-W^* \right \|.
\end{equation}
Considering the local optimal that $\nabla \mathcal{L}(\hat{W}^*, \hat{M})=0$, we have
\begin{equation} 
\label{eq:lambda_c}
\left \| \hat{W}^*-W^* \right \|  \leq \frac{1}{\lambda} \cdot \left \|\nabla \mathcal{L}(W^*, \hat{M}) \right \|.
\end{equation}
Based on Assumption \ref{assumption2}, we get
\begin{equation} \
\label{eq:lambda_c_2}
 \left \| \hat{W}^*-W^* \right \|  \leq \frac{1}{\lambda} \cdot C_{l}.
\end{equation}

For the second term in Eq.\eqref{eq:theta_two_part}, based on Assumption \ref{assumption1}, we have 
\begin{equation} 
\label{eq:}
\begin{aligned}
\left \| \nabla \mathcal{L}(\hat{W}^*, \hat{M})- \nabla \mathcal{L}(\hat{W}^*_e, \hat{M}) \right \| \geq \lambda \cdot \left \| \hat{W}^*-\hat{W}_e^* \right \|.
\end{aligned}
\end{equation}

So, we have 
\begin{equation}
\label{eq:}
\begin{aligned}
\left \| \hat{W}^*-\hat{W}_e^*  \right \| \leq \frac{1}{\lambda} \cdot \left \| \nabla \mathcal{L}(\hat{W}^*, \hat{M})- \nabla \mathcal{L}(\hat{W}^*_e, \hat{M}) \right \|.
\end{aligned}
\end{equation}

Considering the local optimal that $\nabla \mathcal{L}(\hat{W}^*, \hat{M})=0$, we have
\begin{equation} 
\label{eq:delta_w_e}
\begin{aligned}
\left \| \hat{W}^*-\hat{W}_e^* \right \| \leq \frac{1}{\lambda} \cdot \left \| \nabla \mathcal{L}(\hat{W}^*_e, \hat{M}) \right \|=\frac{1}{\lambda} \cdot \left \| \nabla \mathcal{L}(W^*+\Delta W, \hat{M}) \right \|.
\end{aligned}
\end{equation}


Since we use Eq\eqref{eq.prop_1_step4} that $\Delta W = -\frac{\partial L(W^*, \hat{M})^{-1}}{\partial W \partial W} \cdot \frac{\partial L(W^*, \hat{M})}{\partial W}$, we have
\begin{equation}
-\nabla_{W}^2 \mathcal{L}({W}^*, \hat{M}) \cdot \Delta W = \nabla_{W} \mathcal{L}({W}^*, \hat{M})
\end{equation}

So,
\begin{equation} 
\label{eq.long}
\begin{aligned}
&\left \| \nabla \mathcal{L}(\hat{W}^*_e, \hat{M}) \right \|=\left \| \nabla_{W} \mathcal{L}(W^*+\Delta W, \hat{M}) - \nabla_{W} \mathcal{L}({W}^*, \hat{M})-\nabla_{W}^2 \mathcal{L}({W}^*, \hat{M}) \cdot \Delta W \right \|\\
&=\left \|\int_{0}^{1} \left (\nabla_{W} \mathcal{L}^2(W^*+t \cdot \Delta W, \hat{M}) - \nabla_{W} \mathcal{L}^2({W}^*, \hat{M}) \right )  \cdot \Delta W\ dt \right \|\\
&=\left \|\Delta W \cdot \int_{0}^{1} \left (\nabla_{W} \mathcal{L}^2(W^*+t \cdot \Delta W, \hat{M}) - \nabla_{W} \mathcal{L}^2({W}^*, \hat{M}) \right ) \ dt \right \|\\
&\leq \left \| \Delta W \right \| \cdot \left \|\int_{0}^{1} C_{H} \cdot t \cdot \Delta W \ dt \right \|\\
&= \frac{C_H}{2}\left\| \Delta W\right\|^2 \\
&= \frac{C_H}{2}\left \|\frac{\partial^2 \mathcal{L}({W}^*, \hat{M})}{\partial W \partial W}^{-1} \cdot \frac{\partial \mathcal{L}({W}^*, \hat{M})}{\partial W}\right \|^2\\
&\leq \frac{C_H}{2\sigma_{min}^2}\left \|\frac{\partial \mathcal{L}({W}^*, \hat{M})}{\partial W}\right \|^2\\
&\leq \frac{C_H \cdot C_l^2}{2\sigma_{min}^2},\\
\end{aligned}
\end{equation}
where $\sigma_{min}$ is the smallest eigenvalue of Hessian matrix $\frac{\partial^2 \mathcal{L}({W}^*, \hat{M})}{\partial W \partial W}$. The second row in Eq.\eqref{eq.long} is calculated since $\int ( \nabla_{W} \mathcal{L}^2(W^*+t \cdot \Delta W, \hat{M}) \cdot \Delta W) \ dt = \nabla_{W} \mathcal{L}(W^*+t \cdot \Delta W, \hat{M})$. The fourth row is calculated based on Assumption \ref{assumption1} where we assume $\mathcal{L}$ is convex and twice-differentiable with $C_H$.  The sixth row is based on Eq.\eqref{eq.prop_1_step4} that $\Delta W  =-\frac{\partial^2 \mathcal{L}({W}^*, \hat{M})}{\partial W \partial W}^{-1} \cdot \frac{\partial \mathcal{L}({W}^*, \hat{M})}{\partial W}$. The last row is calculated based on Assumption \ref{assumption2}.

Substituting Eq.\eqref{eq.long} into Eq.\eqref{eq:delta_w_e}, we get
\begin{equation}
\label{eq:delta_w_e_2}
\left \| \hat{W}^*-\hat{W}_e^* \right \| \leq \frac{C_H \cdot C_l^2}{2\sigma_{min}^2\lambda}.  
\end{equation}

Based on Eq.\eqref{eq:lambda_c_2} and Eq.\eqref{eq:delta_w_e_2}, $\left \| \Delta W \right \|$ in Eq.\eqref{eq:theta_two_part} can be bounded as:
\begin{equation}
\left \| \Delta W \right \| \leq \frac{1}{\lambda} \cdot C_{l} + \frac{C_H \cdot C_l^2}{2\sigma_{min}^2\lambda}.    
\end{equation}

Then the error bound between the approximated validation loss $L(\hat{W}^*,\hat{M})$ and the ground-truth $\tilde{\mathcal{L}}(\hat{W}^*, \hat{M})$ in Eq.\eqref{eq:error_bound} can be calculated as:
\begin{equation}
E \leq \frac{\left | \Delta W \right |^3}{6} \textup{max}\left | \frac{\partial \mathcal{L}^3}{\partial W^3} \right |,
\end{equation} 
where $\left \| \Delta W \right \| \leq \frac{1}{\lambda} \cdot C_{l} + \frac{C_H \cdot C_l^2}{2\sigma_{min}^2\lambda}$.

\end{proof_c1}

\begin{corollary}
\label{corollary2}
Based on the Assumption \ref{assumption1}-\ref{assumption4}, and when we assign a continuous perturbation on the mask $M$, we could more tightly bound the error between the approximated validation loss $\mathcal{L}(\hat{W}^*, \hat{M})$ and the ground-truth $\tilde{\mathcal{L}}(\hat{W}^*, \hat{M})$ with $E=\left \|\mathcal{L}(\hat{W}^*, \hat{M}) -\tilde{\mathcal{L}}(\hat{W}^*, \hat{M})  \right \| \leq  \frac{K^3}{6} \textup{max}\left | \frac{\partial \mathcal{L}^3}{\partial W^3} \right |$,
where $K=\frac{C_L}{\lambda}  \left \|\Delta M \right \| + \frac{C_HC_a^2}{2\sigma_{min}^2\lambda} \left \| \Delta M\right \|^2+ o(\left \| \Delta M\right \|^2)$.
\end{corollary}
\newenvironment{proof_c2}{\noindent\textbf{\textit{Proof of Corollary \ref{corollary2}:}}}{  $\square$}
\begin{proof_c2}
To get a tighter error bound of approximation in Proposition \ref{prop1}, we need the following more assumptions  (\cite{couellan2016convergence,zhang2021idarts}). 
\begin{assumption}
\label{assumption3}
For any $W$ and $M$, $\mathcal{L}(\cdot, M)$ and $\mathcal{L}(W, \cdot)$ are Lipschitz continuous with $C_f>0$ and $C_L>0$, respectively.
\end{assumption}
\begin{assumption}
\label{assumption4}
$\left\|\frac{\partial^2 \mathcal{L}({W}^*, M)}{\partial M \partial W}\right\|$ is bounded by $C_{a}>0$.    
\end{assumption}

On one hand, based on Assumption \ref{assumption3}, we consider that $L(W^*,M)$ is $C_{L}$ Lipschitz continuous with $M$ and can have
\begin{equation}
\label{eq:corollary_2_step1}
\left \| \nabla \mathcal{L}(\hat{W}^*, \hat{M})- \nabla \mathcal{L}(W^*, \hat{M}) \right \| \leq C_{L} \cdot \left \| \Delta M \right \|.    
\end{equation}

Based on Eq.\eqref{eq.first_term} and Eq.\eqref{eq:corollary_2_step1}, we have
\begin{equation}
\label{eq:corollary_2_step2}
\left \| \hat{W}^*-W^* \right \| \leq \frac{C_{L}}{\lambda} \cdot \left \| \Delta M \right \|.    
\end{equation}

On the other hand, considering a Taylor expansion on $M$ and the local optimal that $\frac{\partial \mathcal{L}({W}^*, {M})}{\partial W}=0$, we have
\begin{equation} 
\label{eq:corollary2}
\begin{aligned}
\left \|\frac{\partial \mathcal{L}({W}^*, \hat{M})}{\partial W}\right \|^2&=\left \|\frac{\partial (\mathcal{L}({W}^*, M)+\Delta M \cdot \frac{\partial \mathcal{L}({W}^*, M)}{\partial M}+o(\Delta M))}{\partial W} \right \|^2\\
&\leq\left \|  \Delta M \frac{\partial^2 \mathcal{L}({W}^*, M)}{\partial M \partial W}  \right \|^2 + o(\left \| \Delta M\right \|^2)\\
&\leq C_a^2\left \| \Delta M\right \|^2+ o(\left \| \Delta M\right \|^2),
\end{aligned}
\end{equation}
where the third row is based on Assumption \ref{assumption4}. 

Substituting Eq.\eqref{eq:corollary2} into Eq.\eqref{eq.long}, we have
\begin{equation}
\label{eq:delta_w_e_3}
\left \| \nabla \mathcal{L}(\hat{W}^*_e, \hat{M}) \right \| \leq \frac{C_H \cdot C_a^2}{2\sigma_{min}^2} \cdot \left \| \Delta M \right \|^2+ o(\left \| \Delta M\right \|^2). 
\end{equation}

Substituting Eq.\eqref{eq:delta_w_e_3} into Eq.\eqref{eq:delta_w_e}, we get
\begin{equation}
\label{eq:result_part2}
\left \| \hat{W}^*-\hat{W}_e^* \right \| \leq \frac{C_H \cdot C_a^2}{2\sigma_{min}^2 \cdot \lambda} \cdot \left \| \Delta M\right \|^2+ o(\left \| \Delta M\right \|^2).
\end{equation}

Based on Eq.\eqref{eq:corollary_2_step2} and Eq.\eqref{eq:result_part2}, $\left \| \Delta W \right \|$ in Eq.\eqref{eq:theta_two_part} can be bounded as: 
\begin{equation}
\left \| \Delta W \right \| \leq \frac{C_L}{\lambda} \cdot \left \|\Delta M \right \| + \frac{C_H \cdot C_a^2}{2\sigma_{min}^2 \cdot \lambda} \cdot \left \| \Delta M\right \|^2+ o(\left \| \Delta M\right \|^2)
\end{equation}

Then the tighter error bound between the approximated validation loss $L(\hat{W}^*,\hat{M})$ and the ground-truth $\tilde{\mathcal{L}}(\hat{W}^*, \hat{M})$ can be calculated as:

\begin{equation}
E \leq \frac{K^3}{6} \textup{max}\left | \frac{\partial \mathcal{L}^3}{\partial W^3} \right |,
\end{equation}
where $K=\frac{C_L}{\lambda} \cdot \left \|\Delta M \right \| + \frac{C_H \cdot C_a^2}{2\sigma_{min}^2\lambda} \cdot \left \| \Delta M\right \|^2+ o(\left \| \Delta M\right \|^2)$.

\end{proof_c2}


\ifCLASSOPTIONcompsoc
\else
\fi


\ifCLASSOPTIONcaptionsoff
  \newpage
\fi



{\small
\bibliographystyle{IEEEtranN}
\bibliography{ref}
}